\documentclass[10pt,twocolumn,letterpaper]{article}

\usepackage[pagenumbers]{cvpr} % To force page numbers, e.g. for an arXiv version
\usepackage{siunitx}
\usepackage{tabularx}
\usepackage{bbding}
\usepackage{pifont}
\usepackage{wasysym}
\usepackage[most]{tcolorbox}
\usepackage{amssymb}
\usepackage{makecell}
\usepackage[dvipsnames]{xcolor} % enable more color names, including Teal
\usepackage{microtype}
\usepackage{booktabs} % for professional tables
\usepackage{multirow}
\usepackage{booktabs}
\usepackage[table]{xcolor}
\usepackage{tcolorbox}
\usepackage{float}
\usepackage{relsize}
\usepackage{textcomp}
\usepackage{array}
\usepackage{fontawesome5}

\usepackage{caption}
\newcommand{\tpm}{\mathrel{\text{\textpm}}}
\definecolor{cvprblue}{rgb}{0.21,0.49,0.74}
\definecolor{mygray}{RGB}{245, 245, 245}
\usepackage[pagebackref,breaklinks,colorlinks,allcolors=cvprblue]{hyperref}

\def\paperID{11825} % *** Enter the Paper ID here
\def\confName{CVPR}
\def\confYear{2025}
\definecolor{mygreen}{rgb}{0,0.4,0}
\definecolor{myred}{HTML}{CF182E}
\definecolor{thaoblue}{HTML}{0F52BA}

\title{Relational Visual Similarity}
\author{
Thao Nguyen$^1$, Sicheng Mo$^2$, Krishna Kumar Singh$^{3}$, Yilin Wang$^{3}$, Jing Shi$^{3}$, Nicholas Kolkin$^{3}$ \\
Eli Shechtman$^{3}$, Yong Jae Lee$^{1,3, \dagger}$, Yuheng Li$^{3, \dagger}$\\
$^{1}$University of Wisconsin-Madison, $^{2}$University of California, Los Angeles, $^{3}$Adobe Research \\
{\smaller[0.5]\url{https://thaoshibe.github.io/relsim}}
}

\newcommand\blfootnote[1]{%
  \begingroup
  \renewcommand\thefootnote{}\footnote{#1}%
  \addtocounter{footnote}{-1}%
  \endgroup
}

\makeatletter
\let\@oldmaketitle\@maketitle
\renewcommand{\@maketitle}{%
  \@oldmaketitle
  \vspace{-5mm}
  \centering
  \includegraphics[width=0.97\linewidth]{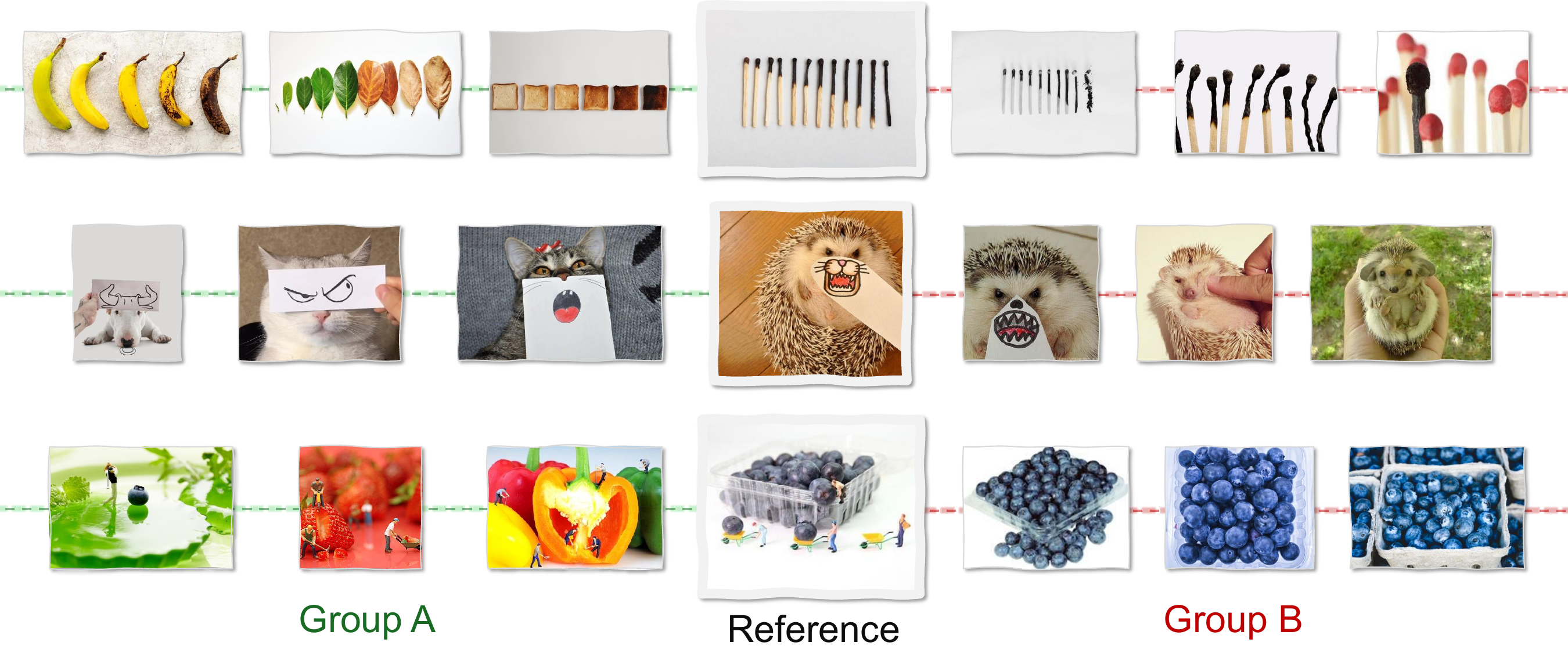}
  \captionof{figure}{
    \textbf{Would you say images in \textcolor{mygreen}{Group A} are similar to the Reference Image?} 
    Current state-of-the-art image similarity models (e.g., LPIPS~\cite{lpips}, CLIP~\cite{clip}) would answer \emph{no}.
    These models would say only \textbf{\textcolor{myred}{Group B}} are similar to the reference image, as they equate similarity with a high degree of shared perceptual attribute features (i.e., color, shape, semantic class).
    However, as humans, we would confidently say \emph{yes}—images in both groups are similar to the reference.
    While \textbf{\textcolor{myred}{Group B}} is similar in perceptual attributes, \textbf{\textcolor{mygreen}{Group A}} is similar in a more abstract, relational sense (e.g., ``transformation of \{subject\} through time'', first row).
    In this paper, we propose to model this missing dimension of visual similarity, or called \emph{relational visual similarity}, capturing human-like reasoning over relational structures.}
  \label{fig:feature-graphic}%
  \vspace{6mm}
}
\makeatother

\begin{document}

\maketitle
\blfootnote{$\dagger$ denotes equal advising}

\begin{abstract}
Humans do not just see attribute similarity---we also see relational similarity. An apple is like a peach because both are reddish fruit, but the Earth is also like a peach: its crust, mantle, and core correspond to the peach's skin, flesh, and pit. This ability to perceive and recognize relational similarity, is arguable by cognitive scientist to be what distinguishes humans from other species. Yet, all widely used visual similarity metrics today (e.g., LPIPS, CLIP, DINO) focus solely on perceptual attribute similarity and fail to capture the rich, often surprising relational similarities that humans perceive. How can we go beyond the visible content of an image to capture its relational properties? How can we bring images with the same relational logic closer together in representation space? To answer these questions, we first formulate relational image similarity as a measurable problem: two images are relationally similar when their internal relations or functions among visual elements correspond, even if their visual attributes differ. We then curate 114k image–caption dataset in which the captions are anonymized---describing the underlying relational logic of the scene rather than its surface content. Using this dataset, we finetune a Vision–Language model to measure the relational similarity between images. This model serves as the first step toward connecting images by their underlying relational structure rather than their visible appearance. Our study shows that while relational similarity has a lot of real-world applications, existing image similarity models fail to capture it---revealing a critical gap in visual computing.

% \vspace{-1mm}
%%%%%%%%% 
%Humans perceive similarity not only through shared attributes but also through shared relations. An apple resembles a peach because both are reddish fruits, yet the Earth also resembles a peach—its crust, mantle, and core correspond to the peach’s skin, flesh, and pit. This capacity for relational similarity—recognizing structural correspondences beyond surface features—is widely regarded as a hallmark of human cognition.However, current visual similarity models (e.g., CLIP, DINO) capture only perceptual or attribute-level similarity, failing to reflect the relational insights that shape human understanding.We propose to rethink visual similarity by modeling relational logic between images. Leveraging vision-language models (VLMs), we connect images through their underlying relational structures and fine-tune representations to align relationally analogous images. Our results reveal a striking gap between human and machine similarity judgments, and demonstrate that relational similarity is both learnable and broadly applicable. This work opens a new direction toward AI systems that reason about visual analogies with human-like insight.
\end{abstract}    
\section{Introduction}
\label{sec:intro}

The ability to perceive and recognize visual similarity is arguably the most fundamental sense for any visual creature, including humans, to interact and make sense of the world~\cite{medin1993respects,hutmacher2019there}.
We process visual attributes to guide decisions: recognizing that a peach is red might signal that it is edible. We also notice similarities across different objects (e.g., shape, color, texture) to categorize, remember, and abstract them: an apple and a peach are both red and round, so they are likely both fruits. Beyond this, we can see relational similarity as well: we abstract familiar patterns to understand more complex or unseen phenomena. For example, we can anticipate the Earth is like a peach, as its layers—crust, mantle, and core—roughly correspond to the peach’s skin, flesh, and pit, even though no one has directly observed it. 
% \yl{add an illustration image may make things more clear} \kr{or use one of the teaser image to explain this}
In cognitive science, attribute similarity and relational similarity are often considered the two central pillars when it comes to understanding human perception of similarity~\cite{medin1990similarity,markman1993structural}.
Attribute similarity underlies everyday activities (e.g., recognition~\cite{shepard1967recognition}, classification~\cite{nosofsky1986attention}, memorization~\cite{tversky1977features}), while relational similarity fuels reasoning and creativity (e.g., analogies~\cite{gentner1983structure}, abstract thought~\cite{gentner1989analogical}). Some researchers argue that relational similarity is even more central to human cognition, as it drives analogical learning and creativity—the traits that set humans apart from other intelligent species~\cite{analogyandsimilarity,holyoak1996mental,gentner2010bootstrapping}.

Unfortunately, current state-of-the-art visual similarity frameworks focus almost exclusively on attribute-level similarity. Traditionally, image similarity in computer vision has been framed as the task of comparing two images and deciding whether they are visually similar, typically at the pixel or feature level using handcrafted descriptors~\cite{sift,dalal2005histograms}.
% or deep learning–extracted embeddings  \yj{cite lpips, dino, clip, etc}.
In recent years, large-scale hierarchical datasets (e.g., ImageNet~\cite{imagenet}) and cross-modal datasets (e.g., LAION-2B~\cite{laion5b}) have enabled deep learning models to move beyond low-level visual details. Modern approaches (e.g.,~\cite{redmon2016you,clip,dino,simonyan2014very,dreamsim,he2016deep}) can recognize different images of the same semantic class or images that match a rough textual description---for example, “a photo of matchsticks”---even if they differ in shape, color, or other low- to mid-level details (Fig.~\ref{fig:feature-graphic}, Group B, first row).

\begin{table}[ht]
\centering
\resizebox{0.99\columnwidth}{!}{
\begin{tabular}{llm{4.8cm}}
\toprule
\textbf{Dataset} & \textbf{Data Format} & \textbf{Example} \\
\midrule
\makecell[l]{BAPPS\\~\cite{lpips}} & \makecell[l]{image triplet\\ \textit{\textcolor{gray}{(low-level perceptual)}}} &
\includegraphics[height=1.16cm, keepaspectratio,page=1]{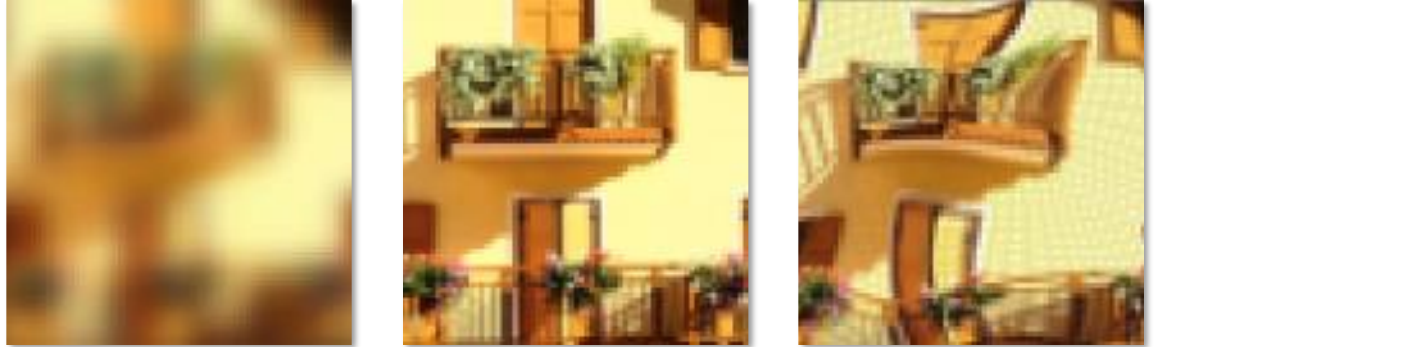} \\
\makecell[l]{NIGHTS\\~\cite{dreamsim}} & \makecell[l]{image triplet\\ \textit{\textcolor{gray}{(mid-level perceptual)}}} &
\includegraphics[height=1.16cm, keepaspectratio,page=2]{final_figures/survey-visual-example-2.pdf} \\
\makecell[l]{ImageNet\\~\cite{imagenet}} & \makecell[l]{semantic class\\ \textit{\textcolor{gray}{(attribute-based)}}} &
\includegraphics[height=1.16cm, keepaspectratio,page=3]{final_figures/survey-visual-example-2.pdf} \\
\makecell[l]{LAION-2B\\~\cite{laion5b}} & \makecell[l]{\{image, caption\}\\ \textit{\textcolor{gray}{(attribute-based)}}} &
\includegraphics[height=1.16cm, keepaspectratio,page=5]{final_figures/survey-visual-example-2.pdf} \\
\midrule
\makecell[l]{\textbf{Ours}\\(\emph{relsim})} & \makecell[l]{\{image, \\ anonymous caption\} \\ \textit{\textcolor{gray}{(relational-based)}} } & \includegraphics[height=1.2cm, keepaspectratio,page=4]{final_figures/survey-visual-example-2.pdf} \\
\bottomrule
\end{tabular}
}
\vspace{-2mm}
\caption{\textbf{Survey of prominent datasets} used for training visual similarity metrics. All are organized based on attribute similarity, whereas ours focuses on relational similarity.}
\vspace{-7mm}
\label{tab:survey}
\end{table}

% However, by focusing primarily on surface-level features, these models struggle to capture \emph{relational} similarity. For instance, they cannot easily recognize that the burning stages of a match resemble the ripening stages of a banana (Fig.~\ref{fig:feature-graphic}, Group B, first row). Capturing this type of similarity requires a shift in perspective: instead of relying solely on visual features, we must reason about how objects, or their parts, interact, abstracting the underlying relationships. For example, both the match and the banana undergo a gradual transformation over time. The similarity lies not in their specific appearance but in the logic of change. This raises fundamental questions: which attributes should be preserved or ignored, and how can we identify, from a single image, which relational patterns are relevant?

However, by focusing primarily on surface-level features, these models struggle to capture \emph{relational similarity} (see ~\cite{Rosenfeld_2018_CVPR_Workshops,4270223}, Sec.~\ref{sec:evaluation}).
% However, by focusing primarily on surface-level features, these models struggle to capture anything beyond \emph{visually look similar} (see Sec.~\ref{sec:evaluation}, also ~\cite{Rosenfeld_2018_CVPR_Workshops,4270223}).
For instance, they cannot easily recognize that the burning stages of a match resemble the ripening stages of a banana (Fig.~\ref{fig:feature-graphic}, Group A, first row). Capturing this type of similarity requires a shift in perspective: instead of relying solely on visual features, we must reason about how different visual elements, interact, abstracting the underlying relationships. For example, both the match and the banana undergo a gradual transformation over time. The similarity lies not in their specific appearance but in the logic of change. This raises questions: which attributes should be preserved or ignored during comparison? How can we identify which relational patterns are relevant or useful?

% \thao{This paragraph need to be rewrite so more motivated like YJ suggestion}
% \kr{I think we should not go into details of using VLM, but it should be more high level like what should model learn, what should be positive pairs}
% Recent advances in Multimodal Large Language Models (VLMs) suggest a promising direction. These models, trained on massive image–text corpora, can understand relational concepts when prompted. However, existing VLMs are not directly suited for relational image similarity at scale, because evaluating relational similarity typically requires pairwise comparisons across an entire dataset. To address this, we propose finetuning a VLM to extract analogous captions—logic-focused descriptions of an image that abstract its relational structure and use placeholders for transferable elements (e.g., “transformation of a \{subject\} over time,” Fig.~\ref{fig:feature-graphic}, first column). By representing each image through its relational logic, we can later finetune the VLM with a contrastive loss to establish the first-ever model for relational image similarity. \thao{I think we need to motivate here more why we go to the text captioning... I will visit this later} \yl{I think we explain and define anonymous caption here }

Insights from cognitive science, encouragingly, offer a spark for these questions. Works~\cite{gentner1983structure,goldstone1994role} showed that humans process attribute similarity perceptually, but relational similarity requires conceptual abstraction, often supported by language or prior knowledge.
This suggests that recognizing relational similarity first requires understanding the image, drawing on knowledge, and abstracting its underlying structure.
Take the example of a photo of burning matches: we first observe how each match relates to the others—they burn sequentially from left to right. With prior knowledge, we understand that burning is a temporal transformation, a process that can occur in many other objects (e.g., a leaf aging, a banana ripening). If asked to write a caption capturing this logic rather than the specific objects, one might write ``transformation of \{subject\} over time''. We call such captions \emph{anonymous captions}---they do not describe any particular visible object but instead capture the relational logic conveyed by the image. These captions act as the glue connecting images with similar underlying logic. In other words, a successful relational visual similarity model must understand, abstract, and use anonymous captions to bring logically similar images together.

To model relational similarity, we follow a path inspired by insights from cognitive science.
Since no existing dataset captures relational visual similarity (see Tab.~\ref{tab:survey}), we first filter a large image corpus, LAION-2B~\cite{laion5b}, to extract 114k images likely to contain transferable relational structures.
% This filtering step ensures that the dataset used for training is of higher quality and more likely to contain meaningful relational relationships that can be transferred to other images, as LAION-2B is known to contain a large fraction of images that are low-quality, mislabeled, or lacking meaningful relational content~\cite{birhane2021multimodal,abbas2023semdedup}.
This step improves dataset quality by removing low-quality, mislabeled, or relationally uninformative images, which are common in LAION-2B~\cite{birhane2021multimodal,abbas2023semdedup}.
We then train an anonymous captioning model to generate captions for these images, creating a set of \{image, anonymous caption\} pairs.
Finally, we train a relational visual similarity model, \emph{relsim}, on this dataset, optimizing it to bring together images whose captions encode similar relational abstractions.
We demonstrate the utility of \emph{relsim} for tasks such as relational image retrieval and analogical image generation.

In short, our contributions are as follows:
\begin{itemize}
\item A new notion of image similarity, \emph{relational visual similarity}, which complements traditional attribute similarity.
\item A novel relational dataset, consisting of 114k \{image-anonymous captions\} designed to capture the abstraction and logic in each image.
\item A new tuned metric, \emph{relsim}, that captures the relational visual similarity between two images.
\item Analysis of the relationship between relational and attribute similarity, along with experiments demonstrating the limitations of current image similarity models.
\item Demonstration of downstream applications in image retrieval and image generation.
\end{itemize}

% This relational framework also enables novel applications. For example, in image generation or editing, a user could provide an image illustrating a relational logic (e.g., a humorous composition combining a person’s photo with a cutout), and the model could generate new images following the same logic. Similarly, relational image retrieval becomes possible: users can search for images not only by attribute similarity but also by deeper relational correspondence—a capability that reflects real-world visual reasoning more closely.
% \thao{REMEMBER TO ADAPT THE CONTRIBUTIONS ACCORDINGLY TO THE EXPERIEMENTS}
% In short, our contributions are as follows:
% \begin{itemize}
% \item We introduce relational similarity as a complement to traditional attribute-based measures, emphasizing structural and functional correspondences between visual elements.
% \item We construct a large-scale dataset of image and their corresponding annomoys caption, focusing on relational similarity.
% \item We propose a metric that evaluates whether visual components interact in analogous ways, going beyond pixel- or feature-level comparisons.
% \item Extensive experiments demonstrate that traditional similarity metrics fail to capture relational correspondences, whereas our approach identifies deeper structural analogies, establishing a benchmark for future work in relational visual reasoning.
% \end{itemize}

\section{Related Works}

% \textbf{Relational similarity.} Relational similarity in cognitive science refers to the correspondence between the relationships among elements in different domains, rather than the similarity of the elements themselves~\cite{Gentner1983}. Humans often reason analogically by aligning relational structures—such as causal, spatial, or functional patterns—while ignoring superficial object features. For example, a solar system and an atom can be considered analogous not because of the objects involved, but because the organizational and relational principles governing their parts are similar. This emphasis on relations over objects provides a foundation for understanding how visual representations can capture higher-order, transferable structures. \thao{this is not checked yet}

\textbf{Similarity in Cognitive Science.} The question of what makes two subjects similar has always been considered one of the most significant questions in cognitive science~\cite{goldstone2012similarity,medin1993respects,tversky2024studies,hahn2013concepts,tversky1977features}. Similarity is fundamental to human cognition, as it affects how the mind organizes, categorizes, and reasons about the world. For decades, Tversky's theory of similarity~\cite{tversky1977features}, also called the contrast model, has been widely adopted and has inspired multiple domains~\cite{lpips,dreamsim,salehi2017tversky}.
Tversky frames similarity as a psychological comparison of matching individual properties or characteristics of objects (e.g., size, shape, color). For example, an apple and a banana are similar because they are both fruits.
While powerful, Tversky's theory cannot account for similarities such as the one Stephen Hawking made when he said, ``I regard the brain as a computer''~\cite{Hawking2011NoHeaven}. There are no obvious visual features shared between a human brain and a computer. This kind of similarity, which cannot be fully accounted for by Tversky's model, was later formalized as relational similarity, alongside its counterpart, now called attribute similarity.
% These concepts emerged from research on analogy, such as Gentner's structure-mapping theory~\cite{gentner1983structure}.
These concepts emerged from Gentner's research on analogy, often referred as Structure-Mapping theory~\cite{gentner1983structure}.
Relational similarity is a comparison based on the relationships between objects. Returning to the previous example, Stephen Hawking was making a relational comparison: he viewed the brain as a biological machine and the process of death as analogous to a computer breaking down. Substantial research shows that while both type of similarity are important, relational similarity (often associated with analogical reasoning) plays a distinct and often deeper role in human cognition (i.e., analogical learning and reasoning ~\cite{medin1993respects,analogyandsimilarity,holyoak1996mental,gentner2010bootstrapping}).
\textbf{Image Similarity.} Comparing similarity between two visual signals is a core concept in computer vision, as it underpins many tasks (e.g., object recognition, image retrieval, image matching). Before the deep learning era, most image similarities were computed directly via pixel-level metrics (e.g., L1, L2, MSE, RMSE, PSNR) or handcrafted features (e.g., SSIM~\cite{ssim}, FSIM~\cite{zhang2011fsim}, SIFT~\cite{sift}). With the rise of deep learning and neural networks (e.g., VGG~\cite{simonyan2014very}, ResNet~\cite{he2016deep}), deep-feature-based image similarity metrics better align with human perceptual judgment (e.g., LPIPS~\cite{lpips}, PieAPP~\cite{prashnani2018pieapp}, DISTS~\cite{DISTS}). More recently, with the aid of Vision Transformers (ViT)~\cite{vit} and Self-Supervised Learning (SSL), modern vision encoders (e.g., DINO~\cite{dino}, CLIP~\cite{clip}, dreamsim~\cite{dreamsim}, SigLIP~\cite{siglip}) not only provide robust visual embeddings for image similarity, but also enable semantic comparisons that go beyond pixel-level matching. However, all of these approaches rely on the assumption that image similarity is based solely on attribute similarity, and thus cannot capture relational similarity, as we demonstrate in our experiments (Sec.~\ref{sec:evaluation}). Here, we, for the first time, propose to consider \emph{relational visual similarity}.

\textbf{Mutimodal Large Language Models.}
Research on multimodal models (e.g., ~\cite{llava,qwenvl25,gpt4o,bagel,xfusion,Nguyen_2025_CVPR,team2023gemini,nguyen2024yo,liu2024improved}) has become an increasingly attractive topic in recent years. In particular, progress in developing unified models that can both understand and generate visual and textual inputs/outputs has transformed how we interpret and interact with visual information. While traditional vision encoders (e.g., CLIP~\cite{clip}) can mostly only ``see'' what is explicitly shown in an image (e.g., ``a photo of a mother hugging a child''), integrating them with MLLMs allows us to capture what is not directly depicted (e.g., ``the image representing a sense of parental care''). Since relational similarity often requires a deeper understanding of images that goes beyond mere perception, we choose to leverage MLLMs, particularly Vision Language Models (VLMs), as the backbone for image feature extraction.
\section{Relational Visual Similarity}

We formalize the problem of measuring the relational visual similarity as follows. Given two input images \( I_{1} \) and \( I_{2} \), we aim to train a visual feature extractor \( f_V \) such that the resulting features capture the \emph{relational similarity} between the two images. Our core assumption is that if two images exhibit high relational similarity, then their corresponding anonymous captions, \( A_{1} \) and \( A_{2} \), should also be similar. 
Specifically, we define the relational similarity score $s_{12}$ between the two images as:
% \vspace{-2mm}
\[
s_{12} = f_V(I_{1}) \cdot f_V(I_{2}) \approx f_{T}(A_{1}) \cdot f_{T}(A_{2}),
\]
% \vspace{-3mm}
where ``\( \cdot \)'' denotes the cosine similarity between the feature embeddings. Here, \( f_T \) represents a textual encoder that produces embeddings for the corresponding captions. 

In Sec.~\ref{sec:create_data}, we describe how to construct the relational dataset, including how to sample image \( \{I_i\}_{i=1}^N \) and generate their corresponding anonymous captions \( \{A_i\}_{i=1}^N \). Then, in Sec.~\ref{sec:train_model}, we detail the training procedure for \( f_V \).

\begin{figure*}[ht] % 'h' means "here" (placement)
    \centering

    \includegraphics[width=0.99\textwidth]{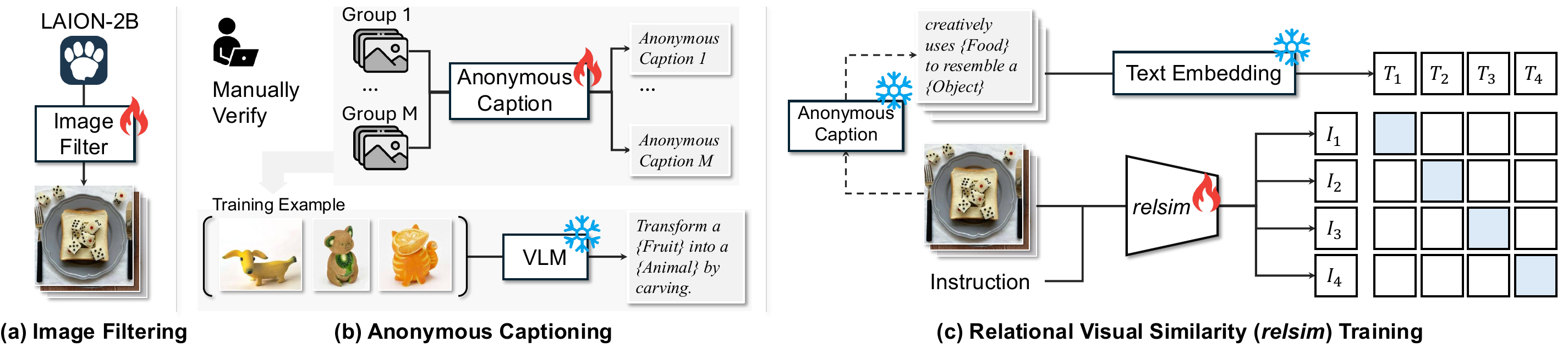}
    \vspace{-2mm}
    \caption{\textbf{Overall pipeline.} (a) We train an image filtering model to select high-quality relational images from LAION-2B~\cite{laion5b}. (b) Anonymous captioning model is trained on groups of images that share the same underlying logic, pairing all images in each group with the same anonymous caption. (c) Training relational visual similarity (\emph{relsim}) model involves a contrastive loss between image features and their corresponding anonymous captions.
    % \thao{Like Krishna's suggestion: Change GPT-4o to VLM?}
    }
    \vspace{-4mm}
    \label{fig:pipeline}
\end{figure*}

\subsection{Creating a Relational Dataset}

\label{sec:create_data}
\begin{figure}
    \centering
    \includegraphics[width=0.9\columnwidth]{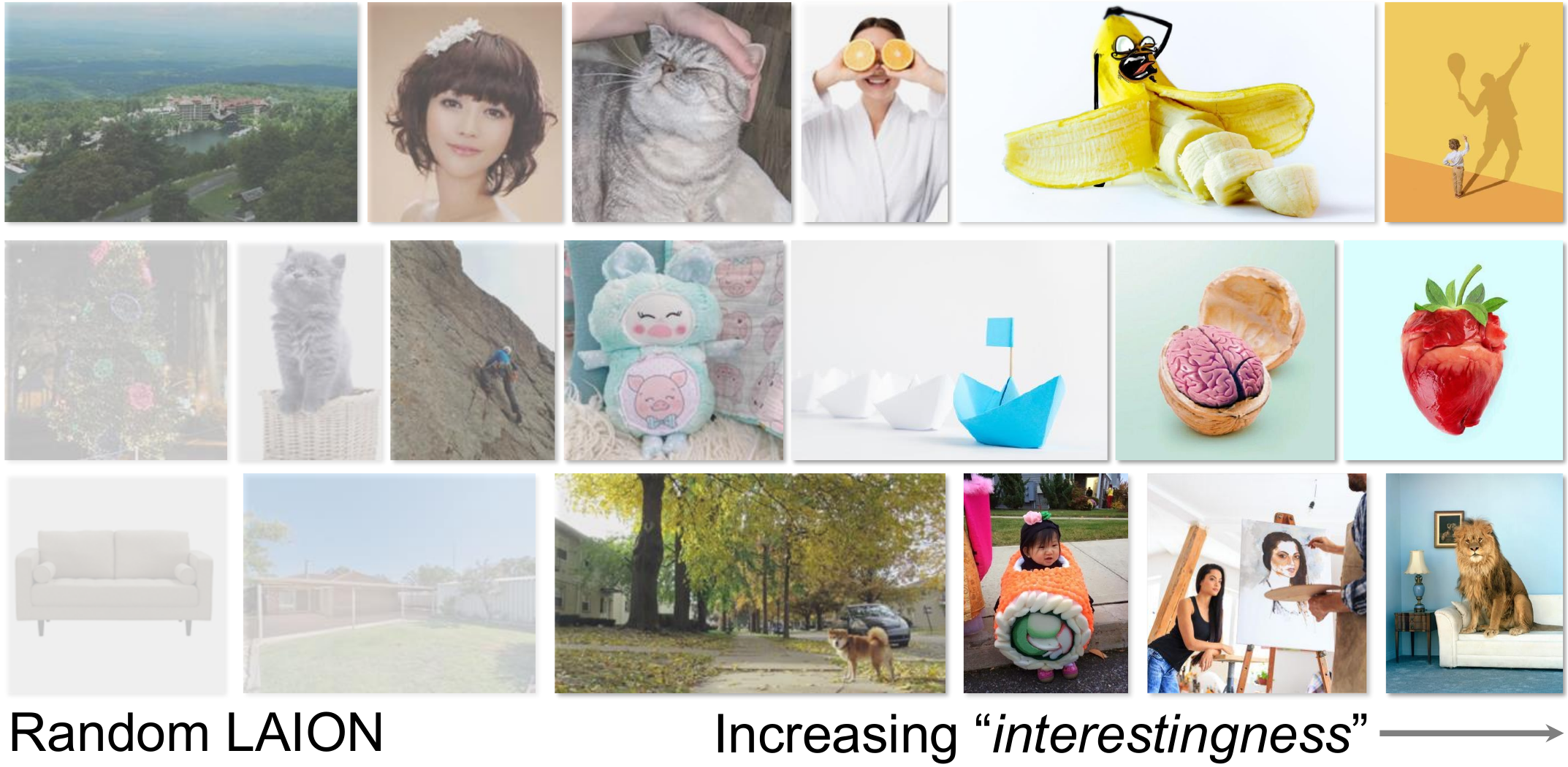}
    % \caption{\textbf{Which images provoke your though about relational attribute?} We argue that not all images are equally relational informative.}
    \vspace{-3mm}
    \caption{Examples of relationally interesting vs. ordinary images.}
    % \caption{\textbf{Relationally interesting vs. ordinary images.} }
    \vspace{-4mm}
    \label{fig:example_interesting_images}
\end{figure}

\textbf{Filtering interesting images $\{I_i\}_{i=1}^N$.} Not all images are equally informative with deep logic for learning relational structures. For instance, an image of a single sofa merely conveys surface-level object appearance, offering limited deep cues about relational organization. In contrast, a photo of ``strawberry heart'' expresses creatively compositional relations that can be abstracted and transferred to new visual content (e.g., ``walnut brain'', Fig.~\ref{fig:example_interesting_images}, second row).

Given the vast nature of LAION-2B, we first perform a filtering step to identify images potentially containing higher-order relational cues (which we refer to as \emph{interesting} images). We fine-tune Qwen2.5-VL-7B-Instruct~\cite{qwenvl25} to classify whether an image is relationally interesting, using 1.3k positive and 11k negative human-labeled examples (Fig.~\ref{fig:pipeline}a). Annotators were instructed: ``Can you see any relational pattern, logic, or structure in this image that could be useful for creating or linking to another image?''.
The fine-tuned model achieves 93\% agreement with human judgments, and when applied to LAION-2B, it yields $N=114k$ images identified as relationally interesting. Details of the prompt and model configuration are provided in the Supp.

\begin{figure}
    \centering
    \includegraphics[width=0.95\linewidth,page=1]{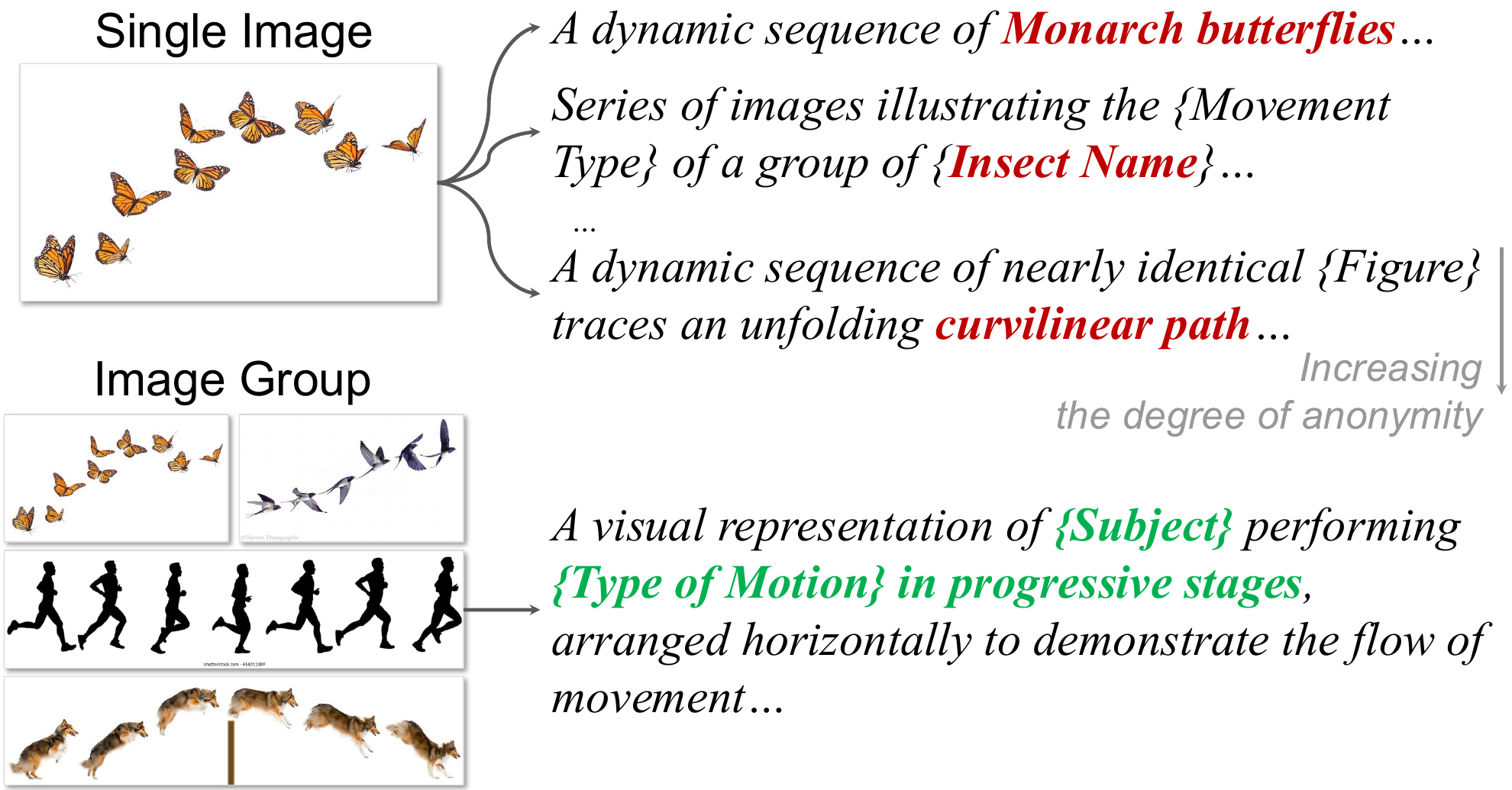}
    \vspace{-2mm}
    % \caption{\thao{Caption... Why we need to caption in the image group} \thao{slightly bigger font size PLEASE}}
    \caption{Writing an anonymous caption is hard from a single image, but easier with an image group where the pattern is clear.}
    \vspace{-4mm}
    \label{fig:caption_by_group}
\end{figure}

\textbf{Generating anonymous captions $\{A_i\}_{i=1}^N$.}
Writing a shared relational attribute from a single image is inherently challenging. For example, given only a sequence depicting a butterfly’s flight stages (Fig.~\ref{fig:caption_by_group}, first row), it is unclear which visual details are irrelevant and which constitute the underlying relational pattern. In contrast, when this image is shown alongside others expressing the same logic (Fig.~\ref{fig:caption_by_group}, second row), the shared relational structure becomes immediately apparent, making it easy to articulate a caption that abstracts away object specifics.

Motivated by this observation, we manually curate $M = 532$ groups of images, where all images within a group exhibit the same underlying relational logic or pattern. Each group has $N_g$ images (a minimum of 2 and a maximum of 10 images). We present each full group to an frozen VLM and prompt it to produce a single anonymous caption $A^{g}$---a relational description that avoids object-specific terms by replacing them with placeholders (e.g., \{subject\}). This caption is then human-verified and paired with every image in the group, yielding an anonymous training dataset (Fig.~\ref{fig:pipeline}b):
% \vspace{-3mm}
\[
\{(I^{g}_i, A^g) \mid i = 1, \dots, N_g\}_{g=1}^{M}
\]
This procedure encourages the model to assign similar anonymous captions to images expressing the same relational pattern. We use  Qwen2.5-VL-7B-Instruct~\cite{qwenvl25} to train this captioning model. After training, we apply it to all ``interesting'' images identified in the previous step, yielding a dataset consisting of images annotated with anonymous relational captions, $\{I_i, A_i\}_{i=1}^N$, where $N = 114,881$ to be exact.

% \thao{explain why we need group with the corresponding figure}

\subsection{Modeling Relational Visual Similarity}
\label{sec:train_model}

% The high level idea is to align the image feature into the logic or abstraction that is represented in the text. \thao{come back here later and add 1-2 sentences about high level plan of training}

% To train the Relational Visual Similarity model (\emph{relsim}), we train a visual extractor $f_V$ that explicitly aligns image representations with their corresponding anonymous captions.

\textbf{Objective.} Given the collection of relationally interesting images with their corresponding anonymous captions $\{(I_i, A_i)\}_{i=1}^{N}$, we train a visual extractor $f_V$ with a frozen text encoder $f_T$ to produce normalized embeddings:
\[
v_i = \frac{f_V(I_i)}{\|f_V(I_i)\|}, \quad
t_i = \frac{f_T(A_i)}{\|f_T(A_i)\|}.
\]
We compute the similarity between an image and its anonymous caption using a dot product scaled by a learnable temperature parameter $\tau > 0$:
\[
s_{ij} = \frac{v_i^\top t_j}{\tau}.
\]
For a batch of size $B$, we use the InfoNCE training loss~\cite{clip}:
% \[
% \mathcal{L}_{\text{CLIP}} = \frac{1}{2B} \sum_{i=1}^{B} 
% \Bigg[
%  -\log \frac{\exp(s_{ii})}{\sum_{j=1}^{B} \exp(s_{ij})}
%  -\log \frac{\exp(s_{ii})}{\sum_{j=1}^{B} \exp(s_{ji})}
% \Bigg].
% \]
\[
\scalebox{1}{$
\mathcal{L} = \frac{1}{B} \sum_{i=1}^{B} 
\Bigg[
 -\log \frac{\exp(s_{ii})}{\sum_{j=1}^{B} \exp(s_{ij})}
 % -\log \frac{\exp(s_{ii})}{\sum_{j=1}^{B} \exp(s_{ji})}
\Bigg]
$}
\]

This training paradigm encourages the visual extractor to capture relationally meaningful features that align with the abstract concepts represented in the anonymous captions.

\textbf{Model Selection.} Traditional visual similarity methods rely on pure vision encoders (e.g.,~\cite{clip,dreamsim,dino}), which derive representations solely from attribute-level features. We find these vision-only encoders insufficient for capturing \emph{relational similarity}, even after tuned, as relational reasoning goes beyond mere visual recognition (See~\ref{sec:evaluation}).

To address this, we leverage Vision Language Models (VLMs) for two reasons: (1) vision encoders emphasize visual attributes or semantics, which can conflict with relational understanding; and (2) relational reasoning often requires higher-level semantic knowledge---which can be found nowhere better than in a Large Language Model, where it was already trained with world knowledge. Accordingly, we employ a VLM as our visual extractor $f_V$ (Fig.~\ref{fig:pipeline}c). Optionally, the task–instruction can be paired with the image as a fixed, steering prompt (e.g., ``Carefully analyze image to understand its underlying logic...'').
\section{Experiments}

\begin{figure*}[t]
    \centering
    % \includegraphics[width=0.99\linewidth,page=1]{figures/baselines.pdf}
    % \caption{Baselines -- Image Retrieval}
    \includegraphics[width=0.98\linewidth]{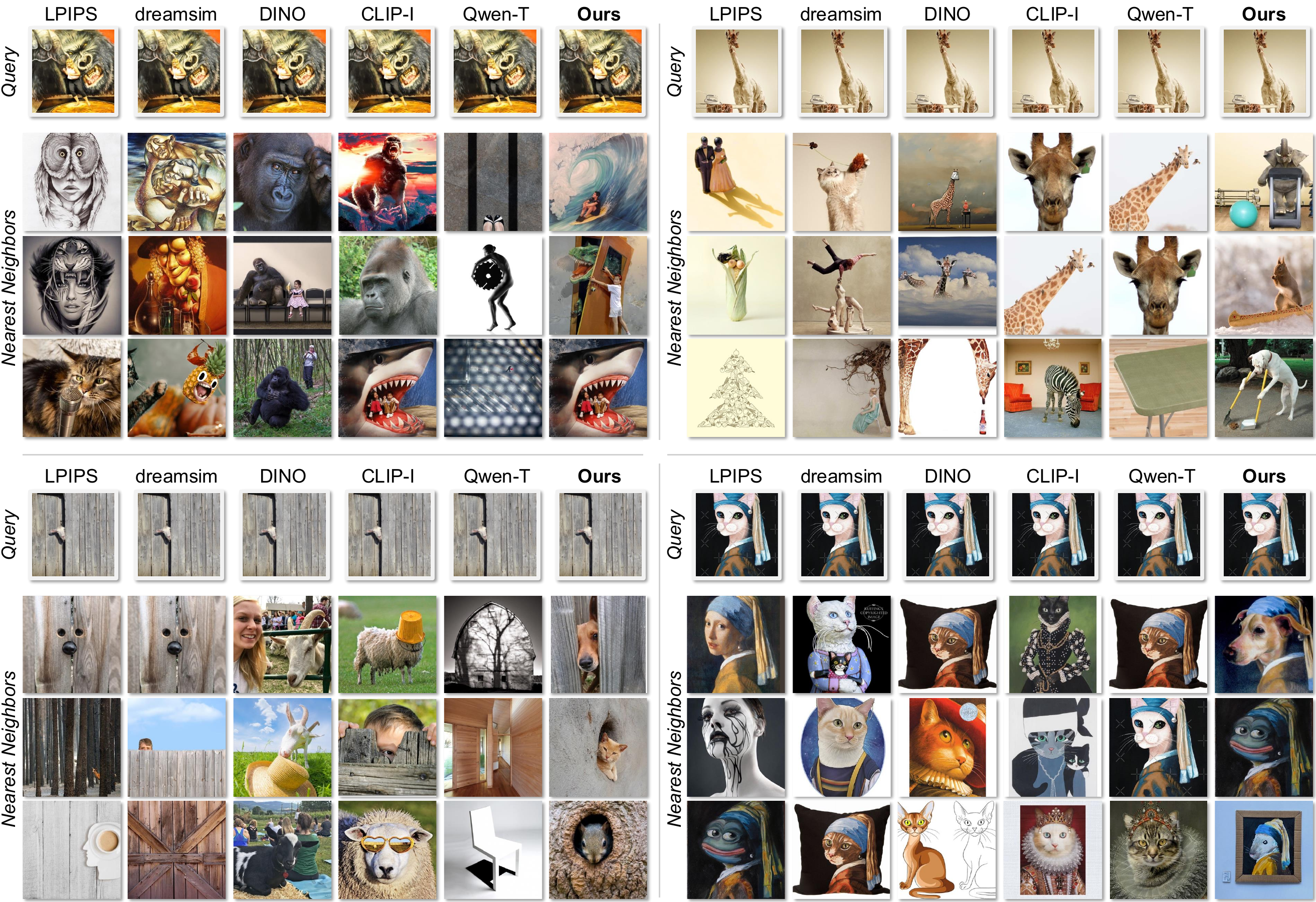}
    \vspace{-2mm}
    \caption{\textbf{Attributes vs. Relational Visual Image Retrieval.} Visualization of nearest neighbor using different visual similarity metrics. As can be seen, only ours understands and can detect the relational similarity.}
    \vspace{-4mm}
    \label{fig:qualitative_baselines}
\end{figure*}

We now discuss our experimental settings, baselines, and evaluation protocol, followed by additional analyses.

\subsection{Settings}
\label{sec:experiment_settings}

\textbf{Implementation.} We adopt Qwen2.5-VL-7B-Instruct~\cite{qwenvl25} as our visual feature extractor $f_V$. Specifically, we append a learnable query token to the end of the image as instruction token, and feed them together into the LLM. We use the query token’s feature from the LLM's last layer as our visual relational feature. For the text embedding model $f_T$, we use all-MiniLM-L6-v2, a widely used and efficient pre-trained model from the Sentence-Transformers library~\cite{sentencetransformer}. We train Qwen2.5-VL-7B-Instruct with LoRA~\cite{lora} for 15k iterations on a single node with 8$\times$A100 GPUs and a batch size of 64.
% See supp for more details. \thao{SUPPLEMENTARY}

\textbf{Data.} To ensure complete separation between training and evaluation, we randomly split the dataset of 114k images into 100k for training and 14k for evaluation.
For evaluation, we consider the image retrieval setting. Specifically, given a query image, we retrieve the most similar image from the database (excluding the query itself); ideally, the retrieved image should be \emph{relationally similar} to the query. The database consists of the 14k images from the test set, combined with another 14k new images randomly sampled from LAION-2B~\cite{laion5b} to better approximate a real-world database.
From this database, 1000 images are randomly chosen from 14k test set to serve as query images.
% From this database, we randomly sample 1,000 images to serve as query images.

% \textbf{Evaluation protocol.} We adopt an MLLM-as-a-judge strategy to evaluate retrieval results. Given a query image and a retrieved image, we prompt GPT-4o~\cite{gpt4o} to rank their relational similarity on a 0–10 scale, where 10 indicates highly relationally similar and 0 indicates no similarity. The full evaluation prompt is provided in the Supplementary. Along with this automatic judegment, we also conduct an user study to see user's reference. User are presented query image, and our retrieved image along side with other baselines, once at a time, and asked to select with one is better (option can be A, B, or Same). This user study enable us to see against with the baseline, how many percent the users prefer our retrieved result over the baseline one.
\textbf{Evaluation protocol.} We employ GPT-4o~\cite{gpt4o} as an automated judge to evaluate retrieval results. For each query image and retrieved image pair, GPT-4o is prompted to assign a relational similarity score on a scale from 0 to 10, where 10 indicates highly relationally similar and 0 indicates no similarity (See Supp. for full prompt).
Along with this automatic evaluation, we conduct a user study to capture human preferences. Participants are shown a query image along with two retrieved images: one from ours and one from a baseline method (randomly named as A or B)---and are asked to select which retrieved image is \emph{relationally more similar} to the query (A, B, or Same). For each baseline, we randomly constructed 300 triplets, and each triplet was independently evaluated by at least three users, resulting in approximately 900 responses per baseline. This study allows us to quantify the proportion of cases in which users prefer our retrieval results over the baselines.
% Using an MLLM as a judge has become common in recent works, as such models can assess relational patterns between images that are difficult for standard vision-only metrics to capture. While MLLMs are reliable for judging relational similarity, efficiently retrieving images from large databases and training a VLM to generate effective feature vectors for this task remain challenging.

\textbf{Baselines.} We compare our approach with prominent image similarity metrics, including LPIPS~\cite{lpips}, DINO~\cite{dino}, dreamsim~\cite{dreamsim}, and CLIP-I~\cite{clip} (image-to-image). These models can directly output similarity scores for a pair of images.
We also consider baselines that operate via captions. In these settings, we first prompt Qwen~\cite{qwenvl25} to generate an anonymous or abstract caption for each image, and then perform retrieval using this caption as the query feature. We evaluate two variants: (1) Apply CLIP-based text-to-image retrieval denoted as CLIP-T; and (2) Text-to-text retrieval denoted as Qwen-T.
Note that in both of these caption-based baselines, we use the original Qwen model rather than our finetuned version. This allows us to show the performance of prompting a VLM to produce the anonymous caption from a single image (see Fig.~\ref{fig:caption_by_group}) whereas finetuned model is our method which benefits from a group of images.

\subsection{Evaluations}
\label{sec:evaluation}

\begin{figure}[ht]
    \centering
    \includegraphics[width=0.97\linewidth]{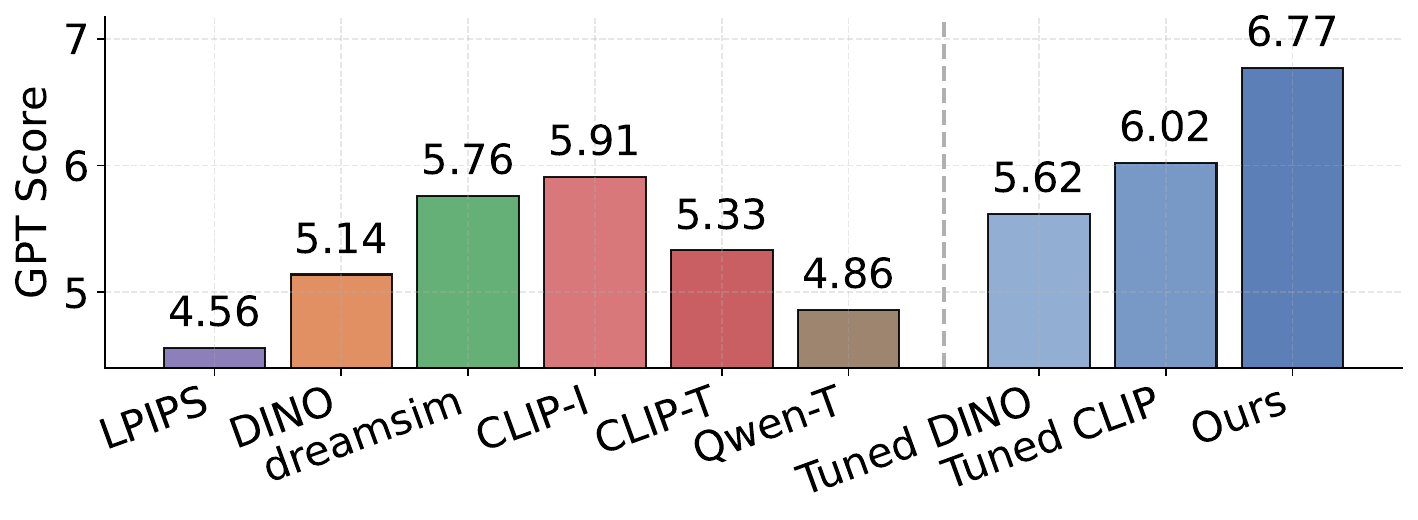}
    %%%%%%%%%%%%$
    % UNCOMMENT THIS IF YOU DON'T LIKE VERTICAL LINE
    % \includegraphics[width=0.98\linewidth]{figures/barchart_academic_color_tuned_no_vertical.pdf}
    \vspace{-4mm}
    \caption{\textbf{Relational visual similarity performance.} All existing image similarity metrics fail to capture relational similarity, even after being tuned. Our final model (\emph{relsim}) which leverages knowledge from VLMs, achieves the highest score (6.77).}
    \vspace{-4mm}
    \label{fig:exp_compare_to_baselines}
\end{figure}

% \begin{figure*}[ht]
%     \centering
%     \includegraphics[width=0.95\linewidth]{figures/xyplot.pdf}
%     \caption{\textbf{Similarity space} showing different kinds of \emph{visual similarity} in terms of degree of relational vs. attribute similarity.}
%     \label{fig:similarity_space}
% \end{figure*}
% \paragraph{Can current image similarity metrics capture relational similarity?} Results are presented in Fig.~\ref{fig:exp_compare_to_baselines}, where higher values indicate better performance (GPT-Score ranges from 0 to 10). As shown, LPIPS, which focuses purely on perceptual similarity, achieves the lowest score (4.56). DINO~\cite{dino} performs only slightly better (5.14), likely because it is trained in a self-supervised manner on image data alone. CLIP-I yields the strongest results among the baselines, which is expected given that CLIP is trained on image–text pairs and its image encoder learns to capture more abstract, semantic information due to image caption. However, CLIP-I still underperforms relative to our method, as our approach explicitly encourages the model to extract even higher-level, more abstract meaning through anonymous captioning. In addition, our vision encoder is equipped with an LLM, enabling deeper reasoning about image.

\textbf{Can existing metrics capture relational similarity?} Results are presented in Fig.~\ref{fig:exp_compare_to_baselines}, where higher values indicate better performance. As shown, LPIPS~\cite{lpips}, which focuses purely on perceptual similarity, achieves the lowest score (4.56). DINO~\cite{dino} performs only slightly better (5.14), likely because it is trained solely in a self-supervised manner on image data. CLIP-I~\cite{clip} yields the strongest results among the baselines (5.91), presumably because some abstraction is sometimes present in image captions. However, CLIP-I still underperforms relative to our method, as achieving a better score may require the ability to reach even higher-level abstractions, such as those in anonymous captions. Our vision encoder, being equipped with LLM knowledge and anonymous captions, yields the highest score (6.77).

% \paragraph{Why we need to generate anonymous caption from a group?} As described in the approach section, our anonymous data are generated from manually selected groups of similar images. Using a group makes it easier to identify the shared relational structure required for a high-quality anonymous caption. The CLIP-T and Qwen-T baselines further illustrate this point: in both cases, anonymous captions are produced from a single image using the original Qwen model~\cite{qwenvl25}. We find that, under this setting, the model is hard to prompt and often leaks semantic or attribute information (see Supp.), causing retrieval to overly focus on semantics rather than relational similarity. In contrast, our finetuned model consistently produces accurate anonymous captions which benefit our relsim model training.

\textbf{Why generate anonymous captions from a group?} As described in the approach section, our anonymous captions are generated from manually selected groups of similar images. Using a group makes it easier to identify the shared relational structure required for a high-quality anonymous caption. The CLIP-T and Qwen-T baselines further illustrate this point (Fig.~\ref{fig:exp_compare_to_baselines}): in both cases, anonymous captions are produced from a single image using the original Qwen2.5-VL-7B-Instruct~\cite{qwenvl25}. We find that, under this setting, the model is hard to prompt and often leaks semantic or attribute information, causing retrieval to overly focus on semantics rather than relational similarity, thus yielding poor results (i.e., 5.33 and 4.86, compared with ours, 6.77).
% Notably, CLIP-Text only reaches 5.33, indicating that generating an anonymous caption from a single image is a non-trivial task, and even advanced VLMs like GPT-4o may still struggle with it. Directly using Qwen does not yield strong results; however, as the base model already has substantial understanding of images and world knowledge, quickly fine-tuning it (Ours) significantly improves performance, achieving a score of 7.23.

% \textbf{Do we really need a VLM?} Our argument is that relational similarity requires more than visual perception—it demands a deeper form of image understanding. Such knowledge is largely absent in vision-encoder-only models.
% To test this hypothesis, we conduct an ablation study in which we finetune several pure vision encoders—CLIP’s vision tower and DINO—using same anonymous captions training data and the same loss. The results, shown in the right panel of Fig.~\ref{fig:exp_compare_to_baselines}, confirm that finetuning with anonymous captions does improve these models’ ability to capture structural relationships. However, their performance still falls short of our full model. This gap is likely because VLMs, which integrate visual features with language-based world knowledge, are inherently better at encoding relational similarity.

\begin{figure*}
    \centering
    \includegraphics[width=0.95\linewidth]{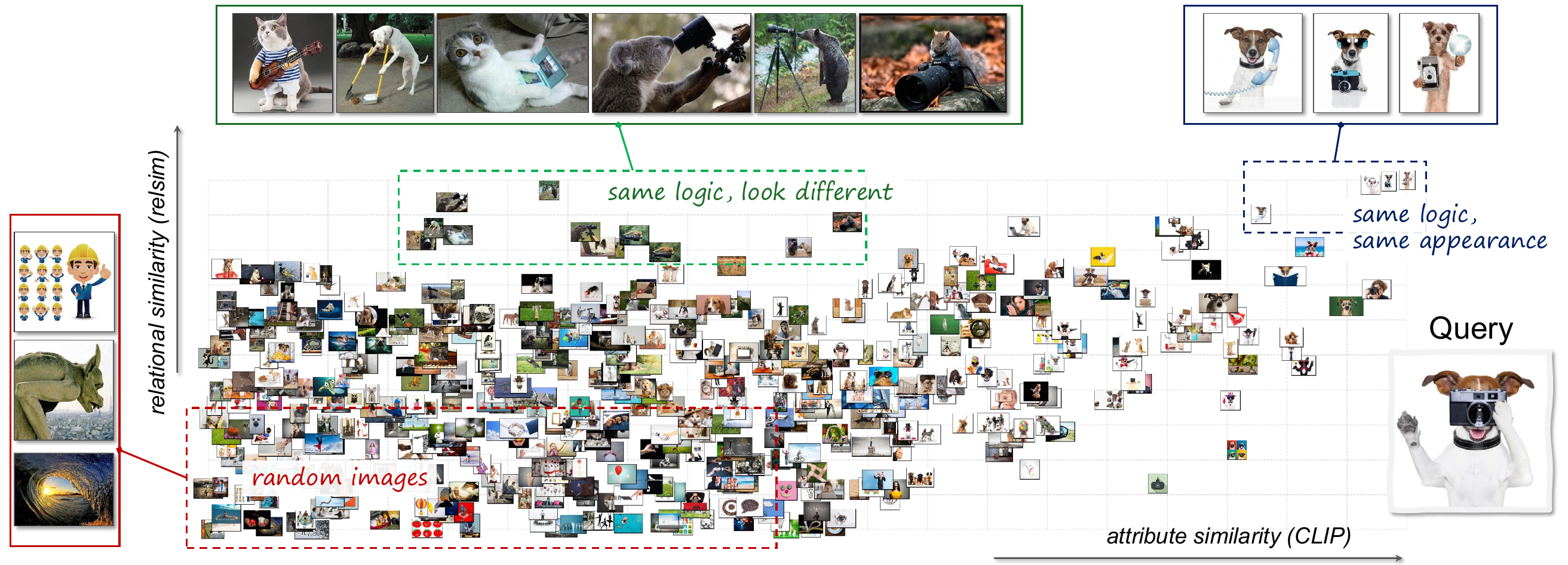}
    \vspace{-3mm}
    \caption{\textbf{Similarity space} showing different kinds of \emph{visual similarity} in terms of degree of relational vs. attribute similarity.}
    \label{fig:similarity_space}
    \vspace{-3mm}
\end{figure*}

\textbf{Knowledge is essential for capturing relational similarity.} Our argument is that relational similarity requires more than visual perception—it demands a deeper form of image understanding. Such knowledge is largely absent in vision-encoder-only models. To test this hypothesis, we conduct an ablation study in which we finetune pure vision encoders (CLIP~\cite{clip} and DINO~\cite{dino}) using the same anonymous captions training data and the same loss. The results (denoted as Tuned CLIP/DINO), shown in the right panel of Fig.~\ref{fig:exp_compare_to_baselines}, indicate that finetuning with anonymous captions does improve these models' ability to capture structural relationships. However, their performance still falls short of our model, which is equipped with a VLM. This gap is likely because VLMs, which integrate visual features with language-based world knowledge, are inherently necessary to understand and encode relational similarity.

\begin{figure}[h]
    \centering
    \vspace{-2mm}
    \includegraphics[width=0.98\linewidth]{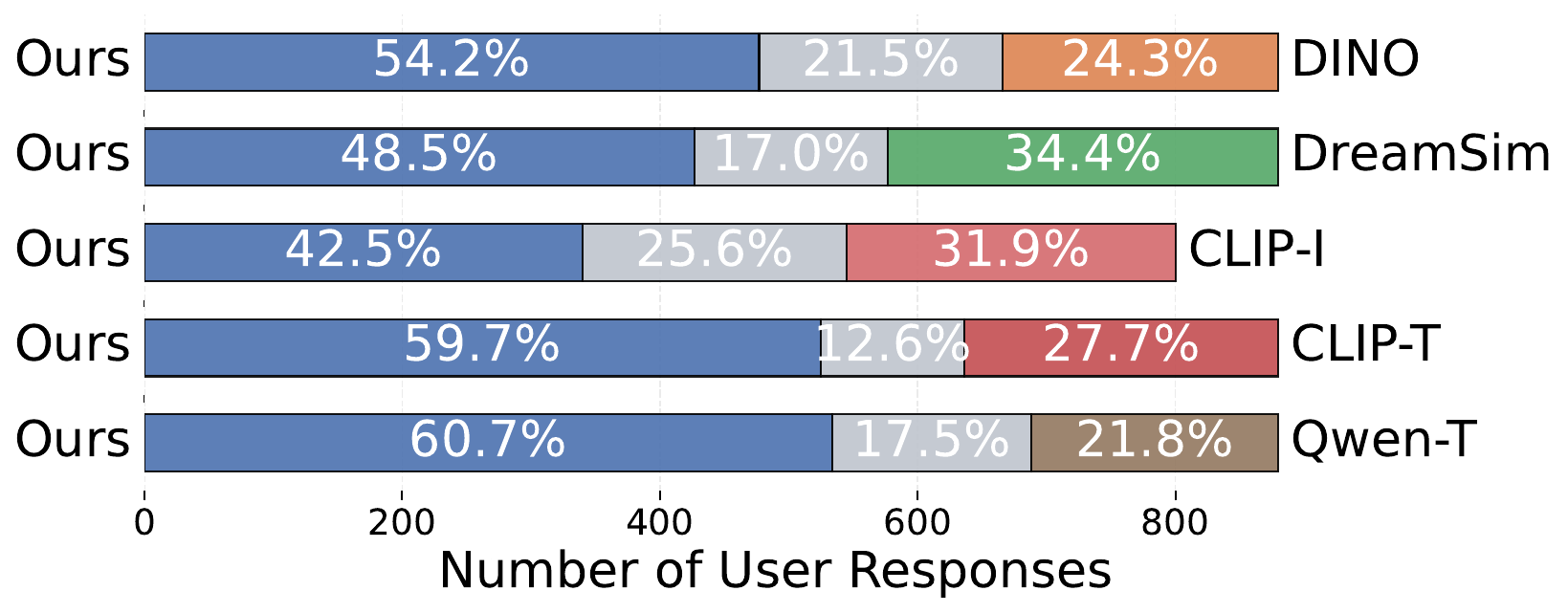}
    \vspace{-2mm}
    \caption{\textbf{User study.} AB testing shows that our model aligns significantly better with human perception of relational similarity compared to the baselines.}% beats all baselines in terms of relational similarity.
    \vspace{-2mm}
    \label{fig:user_study}
\end{figure}

\textbf{Do humans agree with ours?} The result of our user study, shown in Fig.~\ref{fig:user_study}, indicates that users consistently prefer our method across all baseline comparisons, with preference rates ranging from 42.5-60.7\%. The gray bars indicate the tie rate. This is highly encouraging, as it demonstrates not only that our model, \emph{relsim}, can successfully retrieve relationally similar images, but also, again, confirms that humans do perceive relational similarity—not just attribute similarity!

% This effect is especially pronounced in the Qwen-T setting (text-to-text retrieval).
% Similar to the GPT-Score results, CLIP remains the strongest baseline, likely due to its image–text pretraining paradigm, but our method still outperforms it by a clear margin.

\textbf{Relational similarity complements attribute similarity.} At this point, a skeptical reader might ask: then, when to use relational, when to use attribute similarity? The answer is not straightforward. Relational and attribute similarities serve different but complementary roles: while they are often considered separate, combining them can reveal richer structures in visual data. Inspired by the similarity theory~\cite{analogyandsimilarity}, we visualize visual similarity space using a query image ``A dog holding a camera'', and random 3000 images compared to it (Fig.~\ref{fig:similarity_space}). As shown, combining these two aspects of similarity allows us to discover interesting relationships: (1) same logic, same appearance: other photos of similar-looking dogs performing human-like activities; (2) same logic, look different: images of other \{animal\} performing human-like activities; and (3) random images: most other images fall into this category.
This result shows that relational and attribute similarities are, perhaps, most powerful when used together rather than in isolation.

\section{Applications}
% In this section, we demonstrate various scenarios where relational image similarity is particularly useful and can enable powerful applications. These include, but are not limited to, the examples listed below.
In this section, we illustrate scenarios where relational image similarity is useful for downstream applications, including, but not limited to, the examples below.
% \subsection{Similar Idea Generation }
\begin{figure}[ht]
    \centering
    \includegraphics[width=0.98\linewidth]{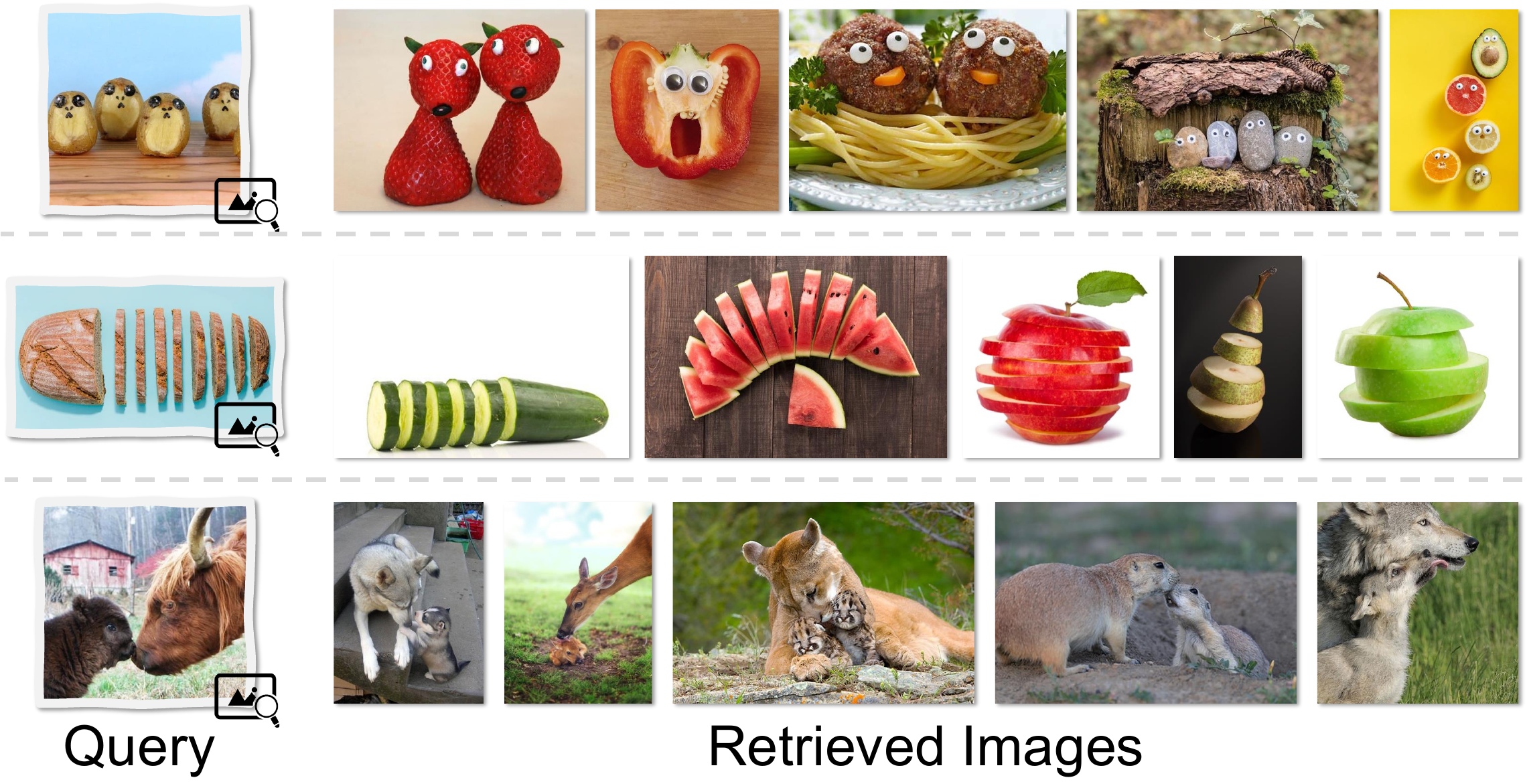}
    \vspace{-2mm}
    \caption{\textbf{Relational image retrieval}. We demonstrate that image can also be searched based on logic or abstraction (relational-based), not only perceptual or semantic similarity.}
    \vspace{-4mm}
    \label{fig:application_image_retrieval}
\end{figure}

\begin{figure*}[ht]
    \centering

    \includegraphics[width=0.98\linewidth]{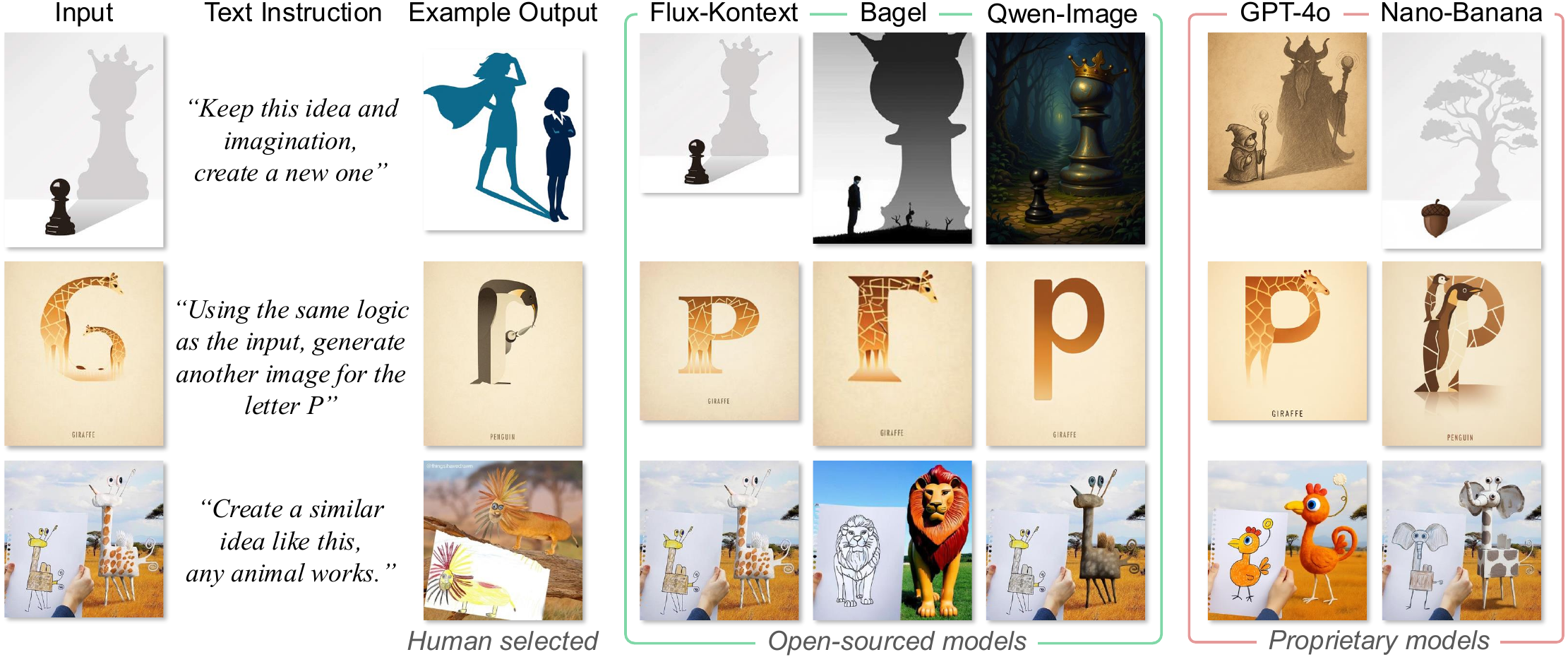}
    \vspace{-2mm}
    \caption{\textbf{Qualitative results for analogical image generation}. Proprietary models are generally better at understanding and performing sophisticated relational transformations, while open-sourced models still lag behind.}
    \vspace{-3mm}
    \label{fig:creative_image_editing}
\end{figure*}

\textbf{Relational image retrieval.} Relational similarity improves retrieval performance in scenarios where attribute-based matching fails, allowing users to search for images not only by semantics but also by higher-level interactions and functions between elements. This approach makes retrieval more aligned with human intuition, which is especially useful for inspiration or creativity. For example, a user might want to retrieve images showing a similarly creative way to decorate a food item with human eyes (Fig.~\ref{fig:application_image_retrieval}, first row).

\begin{figure}[h]
    \centering
    \includegraphics[width=1\linewidth]{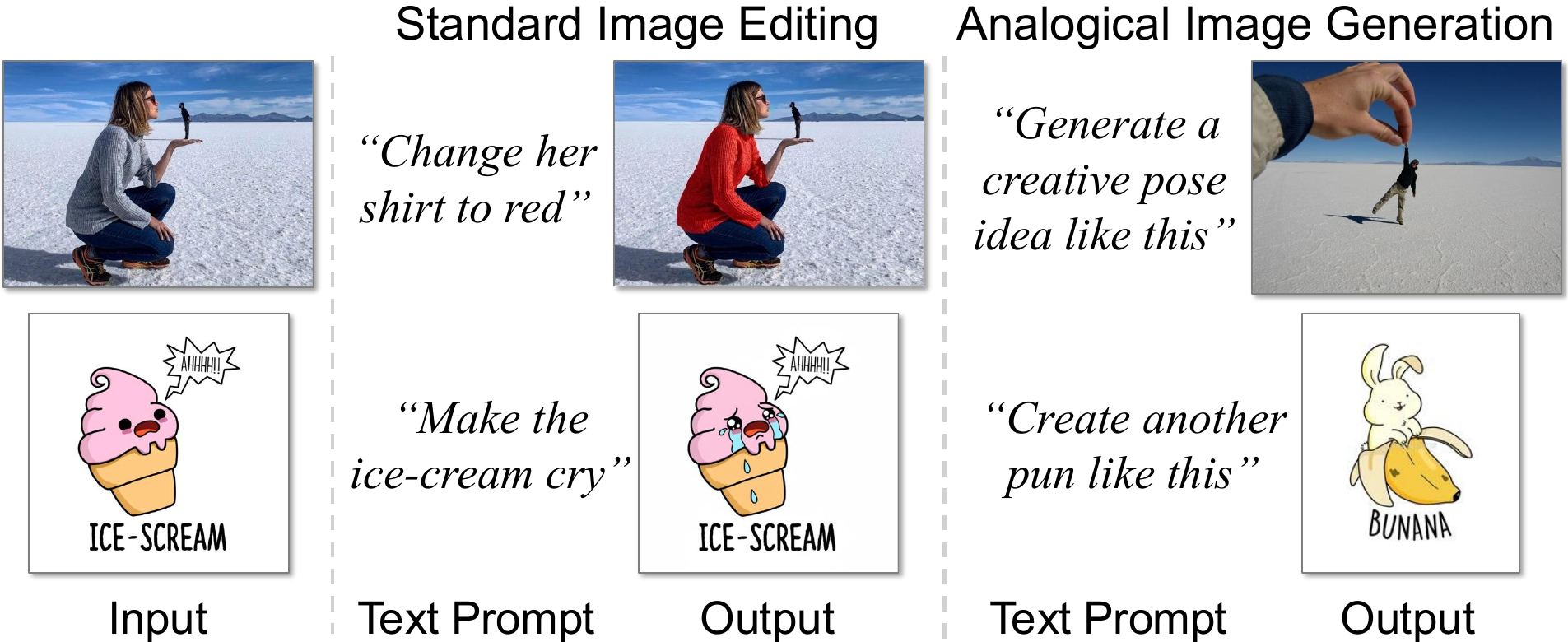}
    % \caption{Standard Editing vs Idea Generation. \thao{aaaa Yuheng can you create a better looking figureeeeee}}
    \vspace{-5mm}
    \caption{\textbf{Analogical image generation}. Unlike standard image editing, which modifies surface attributes, analogical generation transfers deeper relational structures and conceptual ideas.}
    \vspace{-3mm}
    \label{fig:idea_generation}
\end{figure}

\textbf{Analogical image generation.} Relational similarity extends image manipulation beyond surface attributes, allowing the transfer of deeper relational structures and conceptual ideas rather than just shape or texture, unlike conventional image editing. For example, Fig.~\ref{fig:idea_generation} (second row) shows a visual pun realized through typography (i.e., ``ice-scream''); users may wish to generate new images conveying the same concept without predefined constraints on objects or attributes. Evaluating how well current image-editing or MLLM-based methods preserve such relational structures is challenging, but relational similarity provides a promising framework for addressing this gap.

\begin{table}[t]
\centering
\resizebox{0.98\columnwidth}{!}{
\begin{tabular}{lccc}
\toprule
\textbf{Model} & LPIPS ($\downarrow$) & CLIP ($\uparrow$) & relsim ($\uparrow$) \\
\hline
\multicolumn{1}{l}{\textcolor{gray}{\textit{Open-sourced model}}}\\
\hspace{2mm} FLUX-Kontext~\cite{batifol2025flux} & \textbf{0.28 $\tpm$ 0.22} & \textbf{0.87 $\tpm$ 0.12} & 0.71 $\tpm$ 0.26 \\
\hspace{2mm} Bagel~\cite{bagel} & 0.32 $\tpm$ 0.19 & 0.79 $\tpm$ 0.12 & 0.74 $\tpm$ 0.21 \\
\hspace{2mm} Qwen-Image~\cite{wu2025qwen} & 0.29 $\tpm$ 0.21 & 0.86 $\tpm$ 0.13 & 0.71 $\tpm$ 0.22 \\
\multicolumn{1}{l}{\textcolor{gray}{\textit{Proprietary model}}}\\
\hspace{2mm} GPT4o-Image~\cite{gpt4o} & 0.47 $\tpm$ 0.15 & 0.77 $\tpm$ 0.10 & 0.82 $\tpm$ 0.14 \\
\hspace{2mm} Nano-Banana~\cite{comanici2025gemini} & 0.41 $\tpm$ 0.20 & 0.78 $\tpm$ 0.11 & 0.84 $\tpm$ 0.11 \\
\hline
\rowcolor{mygray}
Example Output & 0.60 $\tpm$ 0.17 & 0.66 $\tpm$ 0.11 &  \textbf{0.88 $\tpm$ 0.11} \\
\bottomrule
\end{tabular}
}
\vspace{-2mm}
\caption{\textbf{Quantitative benchmarking of analogical image generation}. LPIPS, CLIP and relsim measure perceptual, semantic, and relational similarity, respectively, between input and edited images.}
\vspace{-5mm}
\label{tab:mlmm_editing_benchmark}
\end{table}

% \thao{Rewrite this part to make it more clear}
% To test this, we built a dataset of 200 image pairs sharing underlying ideas with corresponding transformation instructions (Fig.~\ref{fig:creative_image_editing}) and benchmarked several models. Tab.~\ref{tab:mlmm_editing_benchmark} reports CLIP similarity, LPIPS distance, and relsim similarity to measure semantic, perceptual, and relational structure preservation.
% Results show: (1) \{Input, Output Example\} pairs have low CLIP (0.66), high LPIPS (0.60); and the highest relsim scoE similarity (same logic, look different). (2) Closed-source models (i.e., GPT-4o~\cite{gpt4o}, Nano-Banana~\cite{team2023gemini}) preserve relational logic better, evident by high relsim score (0.8x), while open-source models (i.e., FLUX-Kontext~\cite{batifol2025flux}, Bagel~\cite{bagel},Qwen-Image~\cite{wu2025qwen}) generate visually similar images but often fail to follow the intended transformation, highlighting the need for more challenging relational training data.

To test this, we manually collected 200 image pairs sharing underlying ideas or logic, along with corresponding human-written text instructions, forming triplets: \{``Input'', Text Instruction'', ``Example Output''\} (Fig.~\ref{fig:creative_image_editing}, first three columns). Each triplet reflects a setting where a user provides an input image and a text instruction to generate a new image capturing the same underlying idea or logic.
The results (Tab.~\ref{tab:mlmm_editing_benchmark}) benchmark open-source and proprietary models using CLIP-I~\cite{clip}, LPIPS~\cite{lpips}, and relsim scores to evaluate semantic, perceptual, and relational structure preservation.
Key findings:
(i) Example Outputs can be logically similar to the Input Image (highest relsim: 0.88) while visually differing or belonging to different semantic classes (lowest CLIP: 0.66, highest LPIPS: 0.60), showing that preserving the underlying idea can be more important than visual similarity.
(ii) Open-source models tend to preserve visual similarity (i.e., CLIP: 0.8x) but often miss logical transformations compared to closed-source models (relsim: 0.7x vs. 0.8x) (see Fig.~\ref{fig:creative_image_editing}).
These results highlight both the performance gap between proprietary and closed-source models; and the need for more challenging analogical image generation datasets to improve open-source model training.

\section{Conclusion and Discussion}

% Perceptual similarity enables humans to recognize patterns and group visually similar stimuli, while relational similarity supports higher-level cognition such as analogy, reasoning, and language. Despite its central role in human understanding, relational similarity has been largely overlooked. In this paper, we argue for its importance and show that existing similarity metrics fail to capture it. We introduce a dataset reflecting relational structure, and train a new image similarity metric capable of modeling this aspect. Many applications—including image retrieval and creative image editing—stand to benefit from this richer notion of similarity.

% Our paper is not without limitations. First, because the caption data were manually curated from 532 groups, they may be imperfect and potentially biased; developing methods to automate and scale this process is an important direction for future research. Second, an image can embody multiple relational structures, leading to multiple valid relational mappings. Determining how to use text prompts to specify which relational structure a user intends remains an open question. Nonetheless, by highlighting this overlooked aspect of image similarity, we hope to open new avenues for research in relational understanding for vision systems.

We have proposed \emph{relsim}, a metric modeling \emph{relational visual similarity}---an important aspect of visual understanding that has been largely overlooked. We show that relsim captures image logic and abstraction, which are not effectively measured by existing attribute-based similarity metrics. We further demonstrate several applications of relsim, including visual exploration (image similarity space), image retrieval, and analogical image generation.

That said, our paper is not without limitations. First, the anonymous captioning model is currently trained on 532 manually curated image groups, which may be imperfect, potentially biased, and not scalable. Developing an automated, scalable pipeline to expand these image groups, or relational logics, is an important direction for future research. Second, like other VLMs, the anonymous captioning model can exhibit biases or hallucinations, which can lead to some incorrect captions.
Last but not least, we acknowledge that one image can embody multiple different relational structures, potentially leading to multiple valid relational mappings. Determining how to use text prompts to specify which relational structure a user intends remains an open question.
Nevertheless, our paper highlights relational visual similarity---an overlooked aspect of image similarity---and we hope to open new avenues for future research in relational understanding and generation for vision systems.

\section*{Acknowledgment}
% This work was supported in part by NSF IIS2404180, and Institute of Information \& communications Technology Planning\& Evaluation (IITP) grants funded by the Korea government (MSIT) (No. 2022-0-00871, Development of AI Autonomy and Knowledge Enhancement for AI Agent Collaboration) and (No. RS-2022-00187238, Development of Large Korean Language Model Technology for Efficient Pre-training).

This work was supported in part by NSF IIS2404180, NetApp Inc., and Institute of Information \& communications Technology Planning\& Evaluation (IITP) grants funded by the Korea government (MSIT) (No. 2022-0-00871, Development of AI Autonomy and Knowledge Enhancement for AI Agent Collaboration, (No. RS-2022-00187238, Development of Large Korean Language Model Technology for Efficient Pre-training), and (No. RS-2025-2543949. Environment-Aware and Domain-Adaptive Multimodal Embodied AI for Real-World Interaction).

{
    \small
    \bibliographystyle{unsrt}
    \bibliography{main}

@String(CVPR= {IEEE Conf. Comput. Vis. Pattern Recog.})

@String(ICCV= {Int. Conf. Comput. Vis.})

@String(ICLR = {Int. Conf. Learn. Represent.})

@String(CVPRW= {IEEE Conf. Comput. Vis. Pattern Recog. Worksh.})

@String(CVPR  = {CVPR})

@String(ICCV  = {ICCV})

@String(ICLR  = {ICLR})

@String(CVPRW= {CVPRW})

@InProceedings{4270223,
  author={Shechtman, Eli and Irani, Michal},
  booktitle={CVPR}, 
  title={Matching Local Self-Similarities across Images and Videos}, 
  year={2007},
  }

@InProceedings{Rosenfeld_2018_CVPR_Workshops,
author = {Rosenfeld, Amir and Solbach, Markus D. and Tsotsos, John K.},
title = {Totally Looks Like - How Humans Compare, Compared to Machines},
booktitle = {CVPRw},
year = {2018}
}

@InProceedings{Nguyen_2025_CVPR,
    author    = {Nguyen, Thao and Singh, Krishna Kumar and Shi, Jing and Bui, Trung and Lee, Yong Jae and Li, Yuheng},
    title     = {Yo'Chameleon: Personalized Vision and Language Generation},
    booktitle = {CVPR},
    year      = {2025},
}

@inproceedings{nguyen2024yo,
  title={Yo'llava: Your personalized language and vision assistant},
  author={Nguyen, Thao and Liu, Haotian and Li, Yuheng and Cai, Mu and Ojha, Utkarsh and Lee, Yong Jae},
  booktitle={NeurIPS},
  year={2024}
}

@inproceedings{
  lpips,
  title={The Unreasonable Effectiveness of Deep Features as a Perceptual Metric},
  author={Zhang, Richard and Isola, Phillip and Efros, Alexei A. and Shechtman, Eli and Wang, Oliver},
  booktitle={CVPR},
  year={2018}
}

@inproceedings{dreamsim,
title={DreamSim: Learning New Dimensions of Human Visual Similarity using Synthetic Data},
author= {Fu, Stephanie and Tamir, Netanel and Sundaram, Shobhita and Chai, Lucy and Zhang, Richard and Dekel, Tali and Isola, Phillip},
booktitle={NeurIPS},
year={2023}
}

@inproceedings{imagenet,
  author={Deng, Jia and Dong, Wei and Socher, Richard and Li, Li-Jia and Kai Li and Li Fei-Fei},
  booktitle={CVPR}, 
  title={ImageNet: A large-scale hierarchical image database}, 
  year={2009},
  }

@inproceedings{clip,
  title        = {Learning Transferable Visual Models From Natural Language Supervision},
  author       = {Radford, Alec and Kim, Jong Wook and Hallacy, Chris and Ramesh, Aditya and Goh, Gabriel and Agarwal, Sandhini and Sastry, Girish and Askell, Amanda and Mishkin, Pamela and Clark, Jack and Krueger, Gretchen and Sutskever, Ilya},
  booktitle    = {ICML},
  year         = {2021},
}

@article{qwenvl25,
  title={Qwen2. 5-vl technical report},
  author={Bai, Shuai and Chen, Keqin and Liu, Xuejing and Wang, Jialin and Ge, Wenbin and Song, Sibo and Dang, Kai and Wang, Peng and Wang, Shijie and Tang, Jun and others},
  journal={arXiv},
  year={2025}
}

@inproceedings{xfusion,
  title={X-fusion: Introducing new modality to frozen large language models},
  author={Mo, Sicheng and Nguyen, Thao and Huang, Xun and Iyer, Siddharth Srinivasan and Li, Yijun and Liu, Yuchen and Tandon, Abhishek and Shechtman, Eli and Singh, Krishna Kumar and Lee, Yong Jae and others},
  booktitle={ICCV},
  year={2025}
}

@article{comanici2025gemini,
  title={Gemini 2.5: Pushing the frontier with advanced reasoning, multimodality, long context, and next generation agentic capabilities},
  author={Comanici, Gheorghe and Bieber, Eric and Schaekermann, Mike and Pasupat, Ice and Sachdeva, Noveen and Dhillon, Inderjit and Blistein, Marcel and Ram, Ori and Zhang, Dan and Rosen, Evan and others},
  journal={arXiv},
  year={2025}
}

@article{batifol2025flux,
  title={FLUX.1 Kontext: Flow Matching for In-Context Image Generation and Editing in Latent Space}, 
      author={Black Forest Labs and Stephen Batifol and Andreas Blattmann and Frederic Boesel and Saksham Consul and Cyril Diagne and Tim Dockhorn and Jack English and Zion English and Patrick Esser and Sumith Kulal and Kyle Lacey and Yam Levi and Cheng Li and Dominik Lorenz and Jonas Müller and Dustin Podell and Robin Rombach and Harry Saini and Axel Sauer and Luke Smith},
  journal={arXiv},
  year={2025}
}

@article{wu2025qwen,
  title={Qwen-image technical report},
  author={Wu, Chenfei and Li, Jiahao and Zhou, Jingren and Lin, Junyang and Gao, Kaiyuan and Yan, Kun and Yin, Sheng-ming and Bai, Shuai and Xu, Xiao and Chen, Yilei and others},
  journal={arXiv},
  year={2025}
}

@inproceedings{liu2024improved,
  title={Improved baselines with visual instruction tuning},
  author={Liu, Haotian and Li, Chunyuan and Li, Yuheng and Lee, Yong Jae},
  booktitle={CVPR},
  year={2024}
}

@article{team2023gemini,
  title={Gemini: a family of highly capable multimodal models},
  author={Team, Gemini and Anil, Rohan and Borgeaud, Sebastian and Alayrac, Jean-Baptiste and Yu, Jiahui and Soricut, Radu and Schalkwyk, Johan and Dai, Andrew M and Hauth, Anja and Millican, Katie and others},
journal={arXiv},
  year={2023}
}

@article{bagel,
  title={Emerging properties in unified multimodal pretraining},
  author={Deng, Chaorui and Zhu, Deyao and Li, Kunchang and Gou, Chenhui and Li, Feng and Wang, Zeyu and Zhong, Shu and Yu, Weihao and Nie, Xiaonan and Song, Ziang and others},
journal={arXiv},
  year={2025}
}

@article{hutmacher2019there,
  title={Why is there so much more research on vision than on any other sensory modality?},
  author={Hutmacher, Fabian},
  journal={Frontiers in psychology},
  year={2019},
}

@article{gentner1989analogical,
  title={Analogical learning},
  author={Gentner, Dedre},
  journal={Similarity and analogical reasoning},
  year={1989},
}

@article{markman1993structural,
  title={Structural alignment during similarity comparisons},
  author={Markman, Arthur B and Gentner, Dedre},
  journal={Cognitive psychology},
  year={1993},
}

@inproceedings{lora,
  title={Lora: Low-rank adaptation of large language models.},
  author={Hu, Edward J and Shen, Yelong and Wallis, Phillip and Allen-Zhu, Zeyuan and Li, Yuanzhi and Wang, Shean and Wang, Lu and Chen, Weizhu and others},
  booktitle={ICLR},
  year={2022}
}

@article{sentencetransformer,
  title={Sentence-bert: Sentence embeddings using siamese bert-networks},
  author={Reimers, Nils and Gurevych, Iryna},
  journal={arXiv},
  year={2019}
}

@inproceedings{he2016deep,
  title={Deep residual learning for image recognition},
  author={He, Kaiming and Zhang, Xiangyu and Ren, Shaoqing and Sun, Jian},
  booktitle={CVPR},
  year={2016}
}

@article{llava,
  title={Visual instruction tuning},
  author={Liu, Haotian and Li, Chunyuan and Wu, Qingyang and Lee, Yong Jae},
  journal={NeurIPS},
  year={2023}
}

@article{vit,
  title={An image is worth 16x16 words: Transformers for image recognition at scale},
  author={Dosovitskiy, Alexey},
  journal={arXiv},
  year={2020}
}

@article{abbas2023semdedup,
  title={Semdedup: Data-efficient learning at web-scale through semantic deduplication},
  author={Abbas, Amro and Tirumala, Kushal and Simig, D{\'a}niel and Ganguli, Surya and Morcos, Ari S},
  journal={arXiv},
  year={2023}
}

@article{goldstone2012similarity,
  title={Similarity},
  author={Goldstone, Robert L and Son, Ji Yun},
  year={2012},
journal={The Oxford Handbook of Thinking and Reasoning}
}

@article{birhane2021multimodal,
  title={Multimodal datasets: misogyny, pornography, and malignant stereotypes},
  author={Birhane, Abeba and Prabhu, Vinay Uday and Kahembwe, Emmanuel},
  journal={arXiv},
  year={2021}
}

@article{goldstone1994role,
  title={The role of similarity in categorization: Providing a groundwork},
  author={Goldstone, Robert L},
  journal={Cognition},
  year={1994},
}

@article{simonyan2014very,
  title={Very deep convolutional networks for large-scale image recognition},
  author={Simonyan, Karen and Zisserman, Andrew},
  journal={arXiv},
  year={2014}
}

@article{gpt4o,
  title={Gpt-4o system card},
  author={Hurst, Aaron and Lerer, Adam and Goucher, Adam P and Perelman, Adam and Ramesh, Aditya and Clark, Aidan and Ostrow, AJ and Welihinda, Akila and Hayes, Alan and Radford, Alec and others},
  journal={arXiv},
  year={2024}
}

@inproceedings{redmon2016you,
  title={You only look once: Unified, real-time object detection},
  author={Redmon, Joseph and Divvala, Santosh and Girshick, Ross and Farhadi, Ali},
  booktitle={CVPR},
  year={2016}
}

@inproceedings{dalal2005histograms,
  title={Histograms of oriented gradients for human detection},
  author={Dalal, Navneet and Triggs, Bill},
  booktitle={CVPR},
  year={2005},
}

@article{gentner2010bootstrapping,
  title={Bootstrapping the mind: Analogical processes and symbol systems},
  author={Gentner, Dedre},
  journal={Cognitive science},
  year={2010},
}

@book{holyoak1996mental,
  title={Mental leaps: Analogy in creative thought},
  author={Holyoak, Keith J and Thagard, Paul},
  year={1996},
  publisher={MIT press}
}

@article{analogyandsimilarity,
  title={Structure mapping in analogy and similarity.},
  author={Gentner, Dedre and Markman, Arthur B},
  journal={American psychologist},
  year={1997},
}

@article{gentner1983structure,
  title={Structure-mapping: A theoretical framework for analogy},
  author={Gentner, Dedre},
  journal={Cognitive Science},
  year={1983},
}

@inproceedings{hahn2013concepts,
  title={Concepts and similarity},
  author={Hahn, Ulrike and Chater, Nick},
  booktitle={Knowledge concepts and categories},
  year={2013}
}

@inproceedings{tversky2024studies,
  title={Studies of similarity},
  author={Tversky, Amos and Gati, Itamar},
  booktitle={Cognition and categorization},
  year={2024},
}

@article{medin1993respects,
  title={Respects for similarity.},
  author={Medin, Douglas L and Goldstone, Robert L and Gentner, Dedre},
  journal={Psychological review},
  year={1993},
  publisher={American Psychological Association}
}

@article{medin1990similarity,
  title={Similarity involving attributes and relations: Judgments of similarity and difference are not inverses},
  author={Medin, Douglas L and Goldstone, Robert L and Gentner, Dedre},
  journal={Psychological Science},
  year={1990},
  publisher={SAGE Publications Sage CA: Los Angeles, CA}
}

@article{Hawking2011NoHeaven,
  author       = {Ian Sample},
  title        = {Stephen Hawking: ‘There is no heaven; it’s a fairy story’},
  journal      = {The Guardian},
  year         = {2011},
  url          = {https://www.theguardian.com/science/2011/may/15/stephen-hawking-interview-there-is-no-heaven},
  note         = {Accessed: 2025‑11‑09}
}

@inproceedings{siglip,
  title={Sigmoid loss for language image pre-training},
  author={Zhai, Xiaohua and Mustafa, Basil and Kolesnikov, Alexander and Beyer, Lucas},
  booktitle={CVPR},
  year={2023}
}

@article{nosofsky1986attention,
  title={Attention, similarity, and the identification--categorization relationship.},
  author={Nosofsky, Robert M},
  journal={Journal of experimental psychology: General},
  year={1986},
  publisher={American Psychological Association}
}

@article{shepard1967recognition,
  title={Recognition memory for words, sentences, and pictures},
  author={Shepard, Roger N},
  journal={Journal of verbal Learning and verbal Behavior},
  year={1967},
  publisher={Elsevier}
}

@inproceedings{dino,
  title={Emerging properties in self-supervised vision transformers},
  author={Caron, Mathilde and Touvron, Hugo and Misra, Ishan and J{\'e}gou, Herv{\'e} and Mairal, Julien and Bojanowski, Piotr and Joulin, Armand},
  booktitle={ICCV},
  year={2021}
}

@article{zhang2011fsim,
  title={FSIM: A feature similarity index for image quality assessment},
  author={Zhang, Lin and Zhang, Lei and Mou, Xuanqin and Zhang, David},
  journal={IEEE transactions on Image Processing},
  year={2011},
}

@inproceedings{prashnani2018pieapp,
  title={Pieapp: Perceptual image-error assessment through pairwise preference},
  author={Prashnani, Ekta and Cai, Hong and Mostofi, Yasamin and Sen, Pradeep},
  booktitle={CVPR},
  year={2018}
}

@article{DISTS,
  title={Image Quality Assessment: Unifying Structure and Texture Similarity},
  author={Ding, Keyan and Ma, Kede and Wang, Shiqi and Simoncelli, Eero P.},
  journal = {CoRR},
  year={2020},
}

@article{sift,
  title={Distinctive image features from scale-invariant keypoints},
  author={Lowe, David G},
  journal={International Journal of Computer Vision},
  year={2004},
}

@article{ssim,
  title={Image quality assessment: from error visibility to structural similarity},
  author={Wang, Zhou and Bovik, Alan C and Sheikh, Hamid R and Simoncelli, Eero P},
  journal={IEEE transactions on image processing},
  year={2004},
}

@inproceedings{salehi2017tversky,
  title={Tversky loss function for image segmentation using 3D fully convolutional deep networks},
  author={Salehi, Seyed Sadegh Mohseni and Erdogmus, Deniz and Gholipour, Ali},
  booktitle={International workshop on machine learning in medical imaging},
  year={2017},
}

@article{laion5b,
  title={Laion-5b: An open large-scale dataset for training next generation image-text models},
  author={Schuhmann, Christoph and Beaumont, Romain and Vencu, Richard and Gordon, Cade and Wightman, Ross and Cherti, Mehdi and Coombes, Theo and Katta, Aarush and Mullis, Clayton and Wortsman, Mitchell and others},
  journal={NeuRIPS},
  year={2022}
}

@article{tversky1977features,
  title={Features of similarity.},
  author={Tversky, Amos},
  journal={Psychological review},
  year={1977},
  publisher={American Psychological Association}
}
}

% \clearpage
% \setcounter{page}{1}
\maketitlesupplementary

\begin{figure*}
    \centering
    \includegraphics[width=0.9\linewidth]{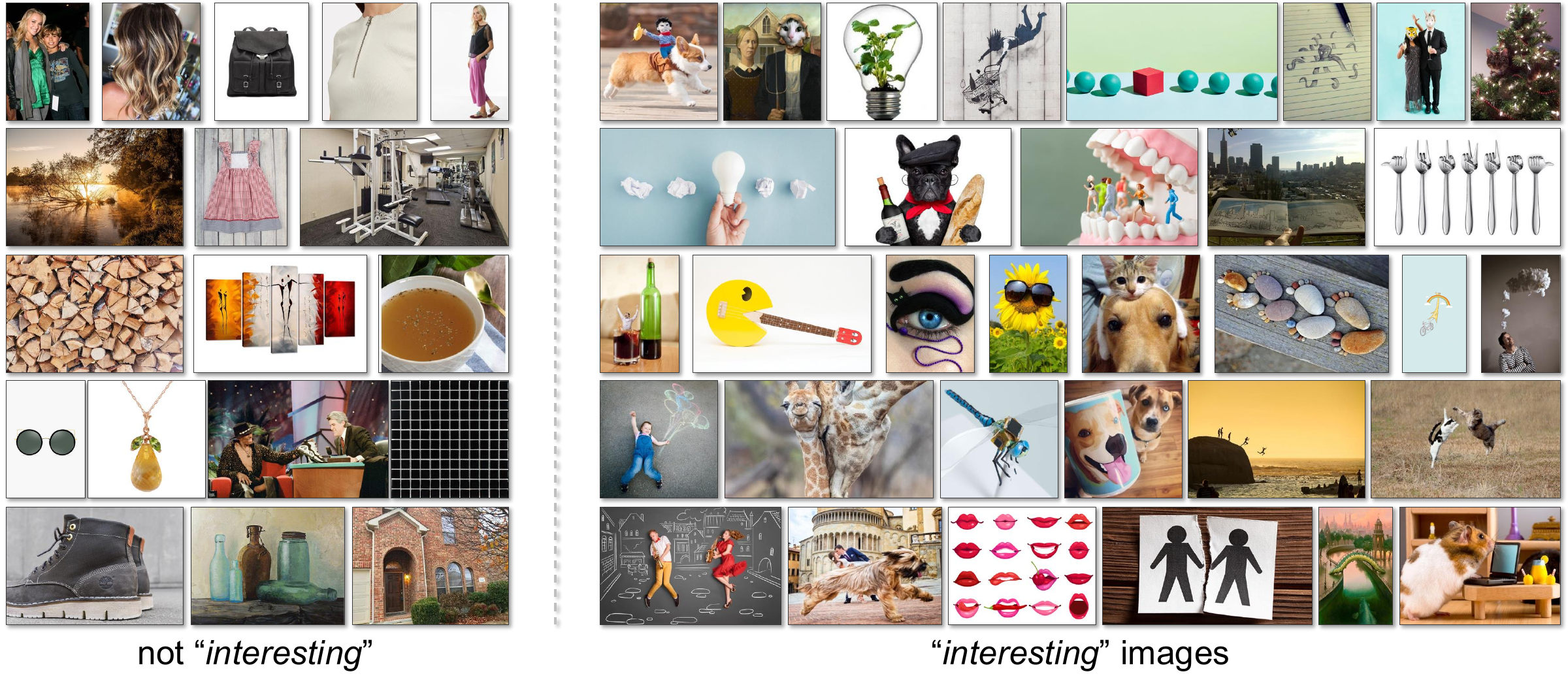}
    \vspace{-3mm}
    \caption{Examples of interesting and uninteresting images filtered by the finetuned Image Filtering model.}
    \label{fig:positive_vs_negative_images}
\end{figure*}

\section{Implementation Details}
\label{supp:implementation_details}
This section presents implementation details as well as snapshots of the training data and model predictions. For specific details about the hyperparameters, etc., please visit our \href{https://github.com/thaoshibe/relsim}{GitHub repository} and the \href{https://thaoshibe.github.io/relsim/data_viewer/index.html}{Hugging Face Datasets page}.
\begin{tcolorbox}[
    title=\textbf{Interesting images filtering prompt},
    coltitle=black,
    colback=gray!10,
    colframe=gray!50,
    arc=2mm
]
You are an expert in visual creativity and interestingness.
Your task is to determine if the given image is visually interesting or not.

If the image is interesting, answer ``Yes''.

If the image is not interesting, answer ``No''.

Remember, you are only allowed to answer ``Yes'' or ``No'', no other words or phrases.
\end{tcolorbox}
\textbf{Interesting Image Filtering.} We trained an image filtering model on 1.3k positive images and 11k negative images. The model used was Qwen2.5-VL-7B-Instruct~\cite{qwenvl25}, trained with LoRA. Positive images were labeled as ``Yes'' (the model should answer ``Yes''), and negative images were labeled  as ``No'' (the model should answer ``No'') accordingly.
Examples of images classified as positive and negative are shown in Fig.~\ref{fig:positive_vs_negative_images}. The keep rate is around 0.7\% (i.e., out of every 1k images, the model marks about 7 as “interesting”).

\begin{tcolorbox}[
    title=\textbf{Write anonymous caption for each image prompt},
    coltitle=black,
    colback=gray!10,
    colframe=gray!50,
    arc=2mm
]
You are given a single image.

Carefully analyze it to understand its underlying logic, layout, structure, or creative concept. Then generate a single, reusable anonymous caption that could describe any image following the same concept.

The caption must:
\begin{itemize}
    \item Fully capture the general logic or analogy of the image.
    \item Include placeholders (e.g., \{Object\}, \{Word\}, \{Character\}, \{Meaning\}, \{Color\}, etc.) wherever variations can occur.
    \item Be concise and standalone.
\end{itemize}

Important: Only output the anonymous caption. Do not provide any explanations or additional text.
\end{tcolorbox}

\begin{figure}[H]
    \centering
    \includegraphics[width=0.95\linewidth]{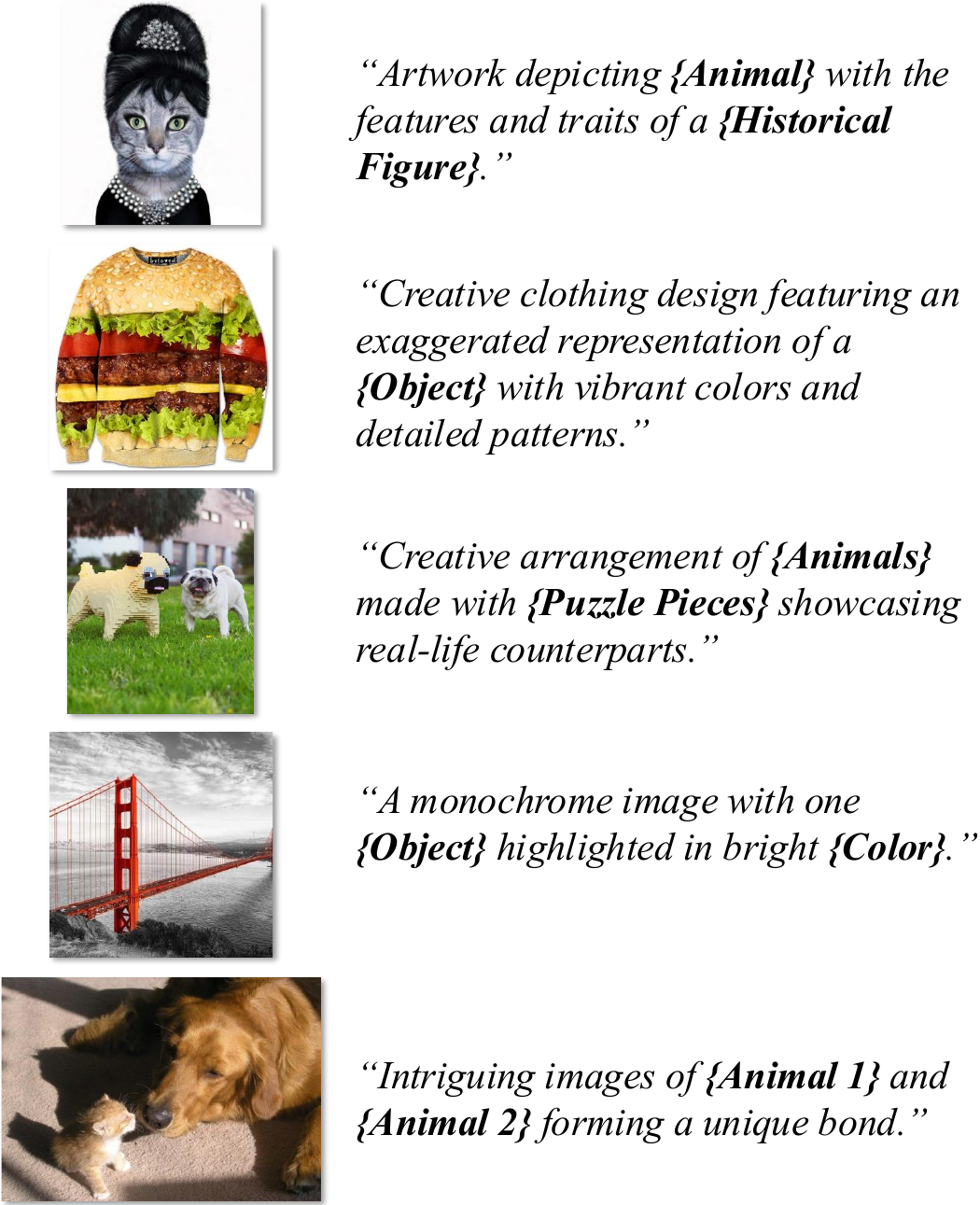}
    \caption{Example of predicted anonymous caption}
    \label{fig:example_of_ano_caption}
    % \vspace{-3mm}
\end{figure}

\textbf{Anonymous captioning model.} The full prompt for obtaining the anonymous captions for each image group, and the prompt used to train the anonymous captioning model, are provided below. We also present an example of a predicted caption for each image in Fig.~\ref{fig:example_of_ano_caption}.

\begin{tcolorbox}[
    title=\textbf{Anonymous captions for image group},
    coltitle=black,
    colback=gray!10,
    colframe=gray!50,
    arc=2mm
]
You are given two or more images that share a common logic, layout, structure, or creative concept (e.g., alphabet worksheets, step-by-step drawings, animals made from peeled fruits, etc.).

Your task is to carefully analyze all the images, identify the shared logic or analogy among them, and create one anonymous caption that describes all the images.

The anonymous caption must:
\begin{itemize}
    \item Be a single, reusable image caption that fully describes the general logic of all the images.
    \item Must include placeholders (e.g., \{Object\}, \{Word\}, \{Character\}, \{Meaning\}, \{Color\}, etc.) wherever variations occur.
\end{itemize}
For example: ``Image of using \{Fruit\} to create a \{Animal\}''; ``Growth process of \{Subject\} described in 4 main stages: \{Stage 1\}, \{Stage 2\}, \{Stage 3\}, \{Stage 4\}''

Only provide the anonymous caption; Do not include any other explanation or content.
\end{tcolorbox}

\textbf{Automated Judgment.} We present the full prompt used for automated judgment of a query image and a retrieved image below.

\begin{tcolorbox}[
    title=\textbf{Automated Judgment for Image Retrieval},
    coltitle=black,
    colback=gray!10,
    colframe=gray!50,
    arc=2mm
]
You are given two images.

Your task is to determine whether these two images share a similar underlying logic---that is, whether they form an analogical pair.

Do NOT base your judgment on visual similarity (e.g., color, shape, composition) or semantic similarity (such as both showing the same object or class). 
Images that are visually or semantically similar but do NOT share the same underlying logic should receive a very low score.

Focus ONLY on whether the two images convey the same conceptual or relational logic. 
For example, if one image shows a peach's internal structures, and the other shows a Earth's internal structures, they share the same logic and should receive a very high score.

Output only the number.
\begin{itemize}
    \item 10 = very strong analogical/relational similarity (same underlying logic)
    \item 0 = no logical/relational similarity
\end{itemize}

Please directly output the score.
\end{tcolorbox}

\section{Additional Results}

\begin{figure*}
    \centering
    \includegraphics[width=1\linewidth,page=1]{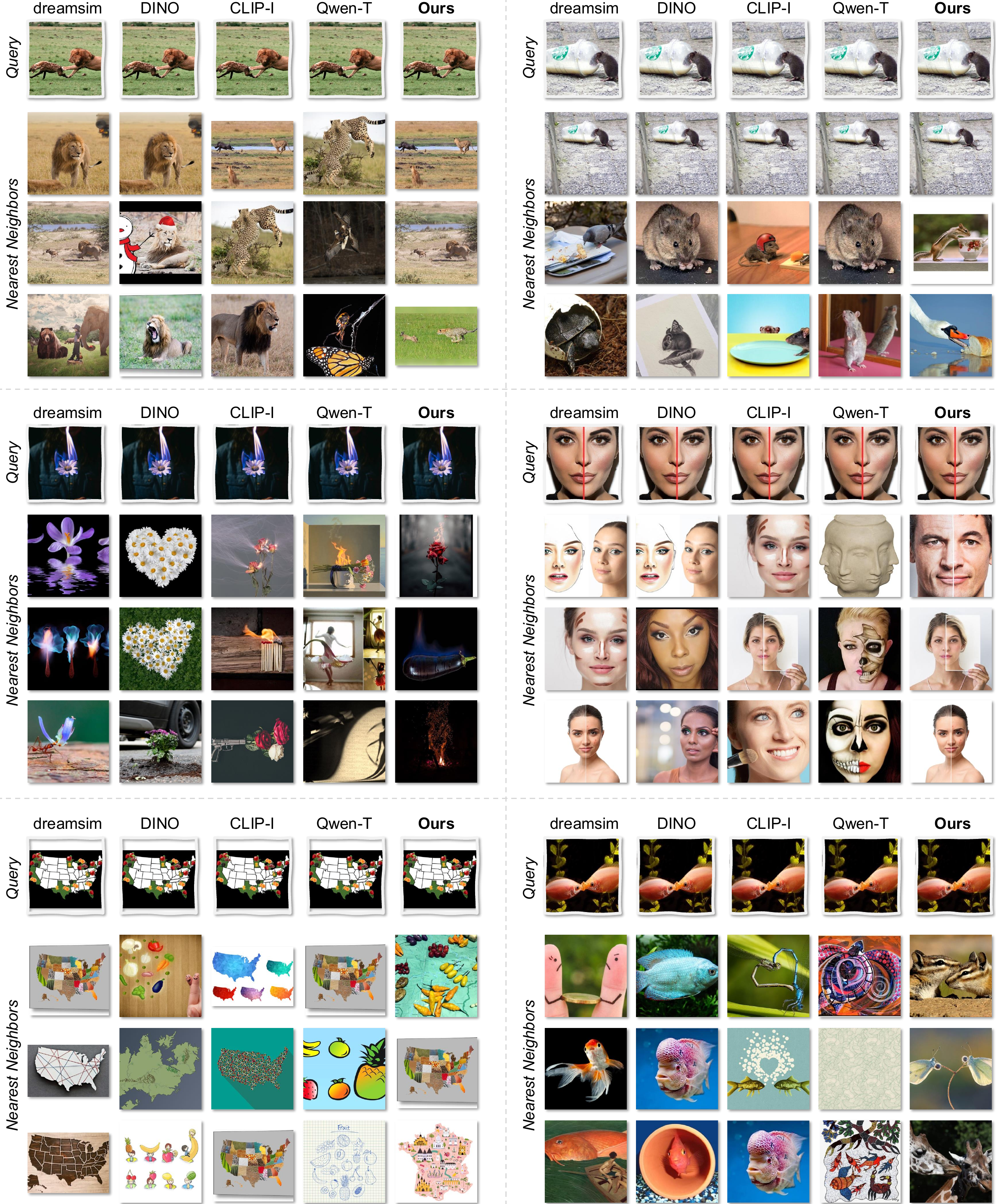}
    \caption{Additional results for image retrieval (1).}
    \label{fig:supp_baseline_1}
\end{figure*}

\begin{figure*}
    \centering
    \includegraphics[width=1\linewidth,page=2]{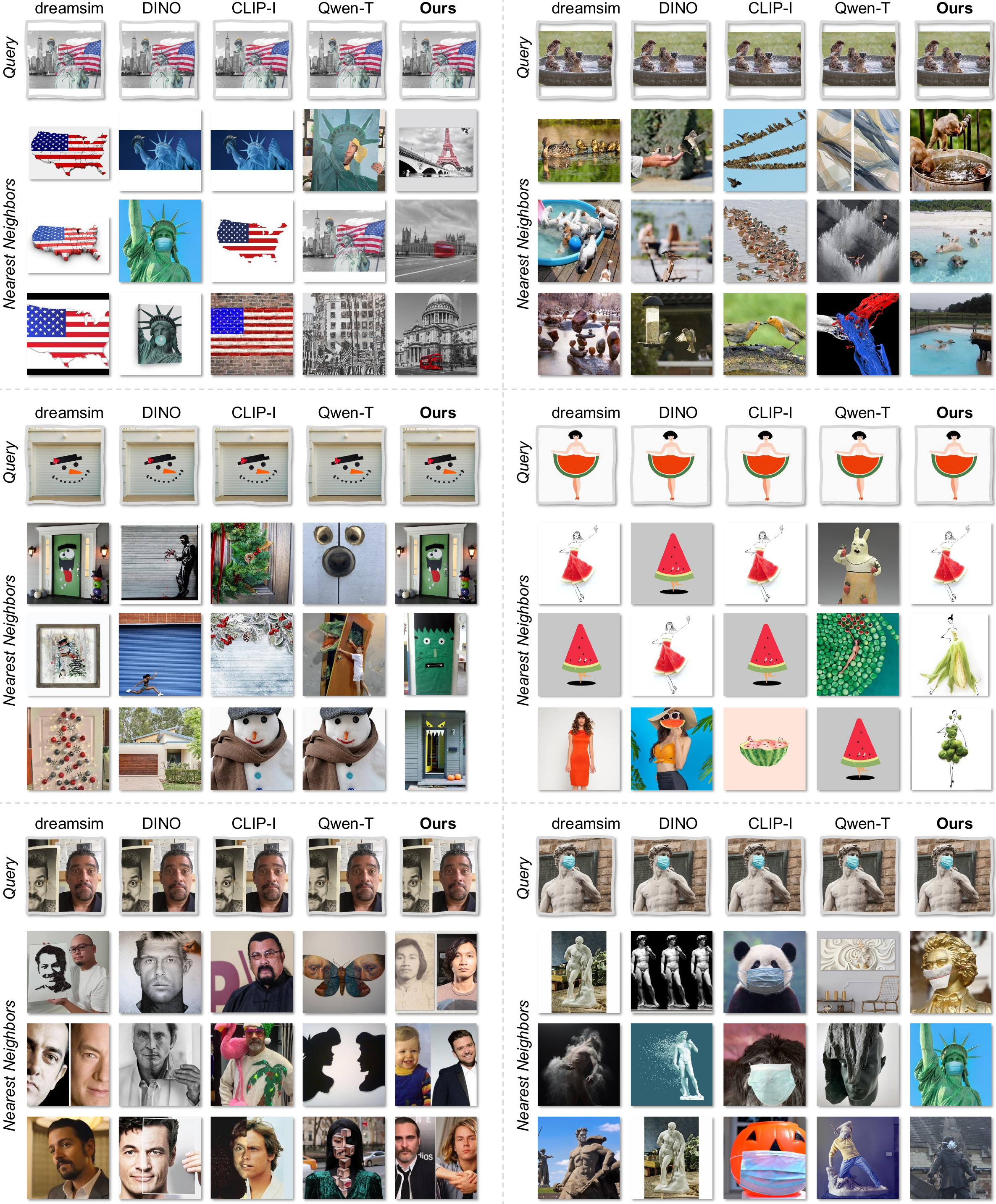}
    \caption{Additional results for image retrieval (2).}
    \label{fig:supp_baseline_2}
\end{figure*}

Additional image retrieval results can be found in Fig.~\ref{fig:supp_baseline_1}-\ref{fig:supp_baseline_2}

\section{Data Collection and Annotation}

The key details of data collection and annotation process are listed as below.

\textbf{About annotators.} All annotation instructions are written in English. All annotators are proficient in English and familiar with Computer Vision (PhD students/holders).

\textbf{Annotating Interesting Images.} Three annotators were shown 10 different groups of ``interesting'' images (e.g., 3 of them are shown in Fig.~\ref{fig:feature-graphic}, Group A) and similarly 5 different groups of ``not interesting'' images (e.g., 3 of them are shown in Fig. Fig.~\ref{fig:feature-graphic}, Group B).
We randomly sampled 15k images from LAION-2B~\cite{laion5b} and simply instructed the annotators to click to select the ``interesting'' images (Line 232). The agreement between these 3 annotators was 92\%.

\textbf{Annotating Image Groups.} 532 image groups have been collected. As above, we showed 10 examples of ``interesting groups'' (e.g., Fig. Fig.~\ref{fig:feature-graphic}, Group A) to nine annotators and asked them to manually find and propose additional groups. In total, $\sim$400 groups were proposed. All proposed groups were further verified by three annotators, and we retained only those for which all annotators agreed that there was a clear, non-duplicate pattern (85\% of the groups was kept).

\section*{Data Attributions}
All images used in this paper are from the publicly available LAION-2B dataset~\cite{laion5b}. The authors do not own any of the images and acknowledge the dataset creators and/or the original copyright holders of each image. All images are used for research purposes only.

\end{document}